\documentclass{article}

    \PassOptionsToPackage{numbers, compress}{natbib}

\usepackage[preprint]{neurips_2025}

\usepackage[utf8]{inputenc} 
\usepackage[T1]{fontenc}    
\usepackage{hyperref}       
\usepackage{url}            
\usepackage{booktabs}       
\usepackage{amsfonts}       
\usepackage{nicefrac}       
\usepackage{microtype}      
\usepackage{xcolor}         

\usepackage{graphicx}
\usepackage{amsmath} 

\usepackage{multirow}
\usepackage{makecell}
\usepackage{colortbl}
\usepackage{color}
\definecolor{lightblue}{RGB}{238, 245, 253}

\usepackage{wasysym}
\usepackage{tcolorbox}
\usepackage{graphicx}
\usepackage{subcaption}  

\usepackage{wrapfig}

\definecolor{modifiedred}{RGB}{0, 102, 204}

\usepackage{graphicx}
\usepackage{float}

\usepackage{titletoc}

\title{LeCoDe: A Benchmark Dataset for Interactive Legal Consultation Dialogue Evaluation}

\author{%
  Weikang Yuan$^{1,2}$\thanks{This work is done when Weikang Yuan works as an intern at Tongyi Lab, Alibaba.}, Kaisong Song$^{2}$, Zhuoren Jiang$^{1}$\thanks{Corresponding author: jiangzhuoren@zju.edu.cn}, Junjie Cao$^{2}$,\\ 
  \textbf{Yujie Zhang}$^{1}$, \textbf{Jun Lin}$^{2}$,  \textbf{Kun Kuang}$^{1}$, \textbf{Ji Zhang}$^{2}$, \textbf{Xiaozhong Liu}$^{3}$\\
  \\
  $^1$ Zhejiang University, China, $^2$ Tongyi Lab, Alibaba Group, China, $^3$ Worcester Polytechnic Institute, USA\\
}

\begin{document}

\maketitle

\begin{abstract}
Legal consultation is essential for safeguarding individual rights and ensuring access to justice, yet remains costly and inaccessible to many individuals due to the shortage of professionals.
While recent advances in Large Language Models (LLMs) offer a promising path toward scalable, low-cost legal assistance, current systems fall short in handling the interactive and knowledge-intensive nature of real-world consultations.
To address these challenges, we introduce LeCoDe, a real-world multi-turn benchmark dataset comprising 3,696 legal consultation dialogues with 110,008 dialogue turns, designed to evaluate and improve LLMs' legal consultation capability.
With LeCoDe, we innovatively collect live-streamed consultations from short-video platforms, providing authentic multi-turn legal consultation dialogues.
The rigorous annotation by legal experts further enhances the dataset with professional insights and expertise.
Furthermore, we propose a comprehensive evaluation framework that assesses LLMs' consultation capabilities in terms of (1) clarification capability and (2) professional advice quality.
This unified framework incorporates 12 metrics across two dimensions.
Through extensive experiments on various general and domain-specific LLMs, our results reveal significant challenges in this task, with even state-of-the-art models like GPT-4 achieving only 39.8\% recall for clarification and 59\% overall score for advice quality, highlighting the complexity of professional consultation scenarios.
Based on these findings, we further explore several strategies to enhance LLMs' legal consultation abilities.
Our benchmark contributes to advancing research in legal domain dialogue systems, particularly in simulating more real-world user-expert interactions.
\end{abstract}

\section{Introduction}
\label{sec:introduction}
Expert consultation services play a vital role in providing professional guidance across knowledge-intensive domains such as law~\citep{xie2024delilaw}, healthcare~\citep{nature2025conversationmedical}, and mental health~\citep{szymanski2025limitations}. 
For example, legal consultation services are crucial for safeguarding individual rights and ensuring fairness in society~\citep{rodrigues2020legal}. 
However, the widening gap between surging legal demands and scarce professional resources has led to prohibitive costs, significantly limiting public who lack strong domain expertise access to justice.
The advance of Large Language Models (LLMs) offers new opportunities to enhance the accessibility and convenience of such services~\citep{expertconsultwork12021precision}, 
providing low-cost legal consultation while potentially reducing lawyers' workloads through their extensive knowledge base and conversational capabilities~\citep{lai2024legalllmsurvey}.

\begin{figure}
    \centering
    \includegraphics[width=0.9\linewidth]{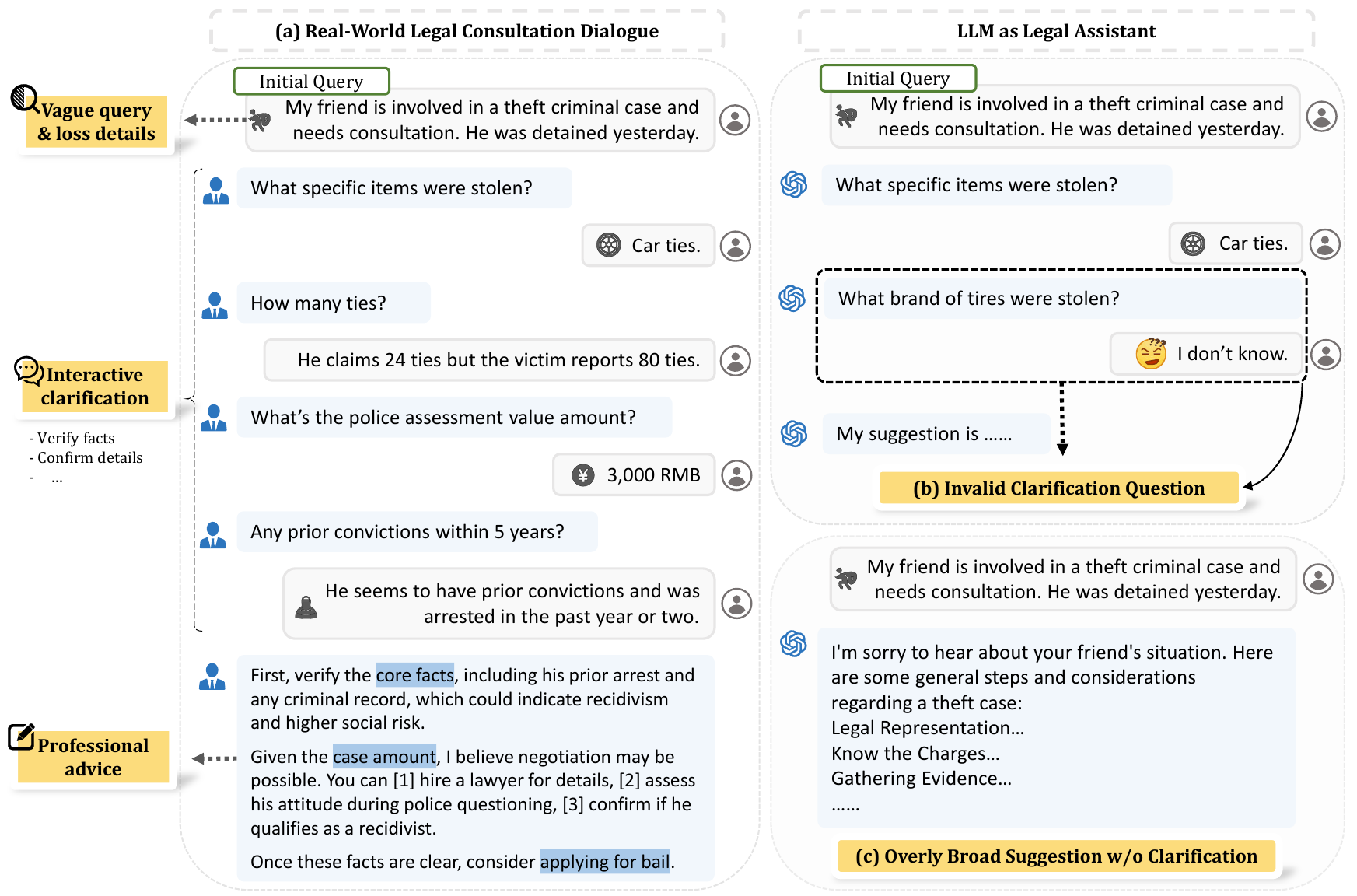}
    \caption{An illustration of real-world legal consultation dialogue.}
    \label{fig1:toyexample}
    \vspace{-5mm}
\end{figure}

As shown in Figure \ref{fig1:toyexample}, the interaction between experts and users in consultation scenarios presents unique complexities that make evaluating LLM-based expert systems particularly challenging~\footnote{While we demonstrate these challenges in the legal domain, similar complexities exist across various consultation scenarios including medical and other professional domains~\citep{fan-etal-2025-ai,li2025beyond}.}.
\textbf{Asymmetric Expertise}: Users who seek for help often lack legal literacy, providing vague initial case descriptions and occasionally omitting critical details.
\textbf{Interactive Consultation Process}: To clarify the users' needs and information, legal experts must employ multi-turn questioning strategies to verify facts and extract case-specific nuances.
Through this iterative clarification, experts integrate both codified knowledge and practical experience to progressively clarify the situation (Clarification Capability) and then deliver professional legal advice (Advice Quality).

Current research on LLM-based expert consultation systems reveals significant limitations. As illustrated in Figure \ref{fig1:toyexample}(b) and \ref{fig1:toyexample}(c), existing models often generate invalid clarification questions or provide immediate advice without necessary clarification interactions.
These limitations can be attributed to two fundamental challenges:
\textbf{Scarcity of Real-world Consultation Data}: Most existing studies rely on single-turn QA pairs~\citep{dai-etal-2025-laiw} or synthetic dialogue data generated by LLMs~\citep{HanFei}. Such datasets inadequately capture the sophisticated dynamics of real-world consultation scenarios, such as users' natural information-seeking behaviors and experts' strategic clarification of missing key information, thus hindering models' ability to learn professional consulting competence.
\textbf{Insufficient Evaluation Framework}: Previous evaluation approaches primarily focus on the quality of final advice~\citep{D3LM} or the accuracy of expert judgments~\citep{li2024mediq}. 
However, this approach overlooks a key aspect: assessing models' clarification ability,  specifically their proficiency in asking purposeful questions before offering advice.

To address these challenges, we introduce the following solutions.
\textbf{First}, to address the scarcity of real-world legal consultation dialogues, we introduce the \underline{Le}gal \underline{Co}nsultation \underline{D}ialogu\underline{e} Dataset (LeCoDe), a large-scale benchmark dataset constructed from real-world legal consultations.
Our research leverages a novel and previously unexplored data source: live-streamed legal Q\&A sessions hosted on Chinese short-video platforms.
In this unique digital environment, licensed lawyers select representative legal cases for real-time consultations, subsequently publishing real-world case dialogues as anonymized educational videos.
This data collection approach ensures authenticity and professional expertise, providing unprecedented access to genuine legal consultation dialogues that existing benchmark datasets rarely capture.

Furthermore, we meticulously design a pipeline to transform these video consultations into high-quality interactive dialogue datasets.
Through careful video processing and dialogue transcription, followed by multi-stage annotations conducted by a team of professional annotators with legal expertise. We further enhanced the dataset by adding richer expert annotations, including utterance intents, key facts with their relative importance, and the summary of expert advice.
This fine-grained annotation not only enables more precise evaluation but also provides valuable guidance for model training.
Finally, our rigorous annotation process yields a knowledge-intensive and structured dataset comprising 3,696 legal consultations with 110,008 dialogue turns.

\textbf{Second}, we propose an interactive legal consultation framework to comprehensively evaluate LLMs' consultation capabilities. 
This framework enables systematic assessment through simulated user-expert interactions, measuring both \textbf{Clarification Capabilities} (in terms of \textit{effectiveness} and \textit{efficiency}) and the \textbf{Advice Quality} through \textit{automated metrics} and \textit{LLM-based evaluation}. 
We conduct extensive experiments on both general LLMs and legal domain-specific LLMs to evaluate their performance in legal consultation scenarios.
Our findings reveal significant limitations in current LLMs' legal consultation capabilities. In terms of clarification ability, even SOTA LLMs like Qwen-max and GPT-4 achieve only modest recall rates of 39.8\% and 35.9\% respectively, indicating their ability to elicit key facts remains insufficient, capturing merely 30-40\% of critical information.
Similarly, in advice quality, leading models like Deepseek and GPT-4 achieve overall scores of only 62\% and 59\%, demonstrating poor performance across profession, completeness, and client satisfaction metrics.
To effectively leverage the LeCoDe dataset, we propose various strategies for constructing SFT training dialogues. While these approaches demonstrate meaningful improvements in model performance, a substantial gap remains between current capabilities and the requirements of real-world legal consultation. 
These insights establish LeCoDe as both a challenging benchmark for evaluating legal AI systems and a valuable resource for advancing their development.

Our key contributions are threefold:

(1) \textbf{High-Quality Dataset from Previously Unexplored Data Source}: We introduce LeCoDe, the first large-scale benchmark dataset explicitly designed for interactive legal consultations, constructed from authentic multi-turn dialogues. By leveraging a novel data source and rigorous expert annotation, LeCoDe provides an unprecedentedly authentic, professionally validated resource.
It significantly enriches evaluation for real-world legal AI applications. To the best of our knowledge, LeCoDe is the first and largest real-world multi-turn legal consultation dialogue dataset.

(2) \textbf{Comprehensive Evaluation Framework}: We establish a comprehensive evaluation framework tailored specifically for legal consultation tasks. Unlike prior methods, our framework systematically evaluates both the critical clarification interactions and the final quality of professional advice through fine-grained metrics. This comprehensive approach enables more precise assessment and targeted improvements of models’ consultation performance.

(3) \textbf{In-depth Empirical Analysis and Strategic Insights}: Our extensive experiments reveal substantial performance gaps among state-of-the-art LLMs when evaluated under realistic legal consultation scenarios. Crucially, we explore and propose several effective Supervised Fine-Tuning (SFT) strategies that demonstrably improve model performance. These empirical insights lay a solid foundation and offer actionable strategies for future advancements in Legal AI research.

\section{Related Work}

Large Language Models (LLMs) have exhibited impressive capabilities and performance across diverse areas~\citep{achiam2023gpt,bai2023qwen,guo2025deepseek}, and their interactive nature with users shows significant potential in consultation settings~\citep{li2025beyond}.
However, a critical challenge emerges in consultation scenarios: users often provide vague queries, requiring LLMs to possess robust clarification abilities to address this information gap~\citep{fan-etal-2025-ai,liu2023cikmask,nature2025conversationmedical,D3LM}.

In the legal domain, existing benchmarks typically evaluate the foundational judicial reasoning capabilities of LLMs, like legalbench~\citep{legalbench} and lexeval~\citep{lexeval}.
Existing works have attempted to enhance LLMs capability in consultation scenarios by supervised fine-tuning~\citep{sun2024lawluo}, reinforcement learning~\citep{D3LM}, and multi-agent collaboration~\citep{cui2023chatlaw}.
However, existing works often overlook the interactive nature of legal consultations and lack access to real-world consultation data.
CrimeKgAssistant contains 200K real-world lawyer-client QA pairs but is limited to single-turn interactions~\citep{dai-etal-2025-laiw}.
Hanfei~\citep{HanFei} provides multi-turn dialogues but relies on synthetic conversations generated by LLMs.
CAIL2023 conversational similar case retrieval dataset (CAIL2023-ConvIR)~\citep{cail2023} focuses on retrieval task without providing legal advice or real scenarios, and CAIL2024 consultation dialogue generation dataset (CAIL2024-ConGen)~\citep{cail2024} utilizes generated rather than real-world data.

\begin{table}[h]
\vspace{-3mm}
  \caption{Related Datasets Comparison: According to Presence of Legal \textbf{Advice}, Inclusion of \textbf{Clarifying} Questions,\textbf{ Multi-Turn} Dialogue Capability, \textbf{Expert Ann}otation Status, and Data \textbf{Source} (LLM-\textbf{Gen}erated vs. \textbf{Real}-world Scenario Data)
}
  \label{tbl:relatedwork}
  \centering
  \scalebox{0.85}{
  \begin{tabular}{l|ccccc}
    \toprule
\textbf{Datasets}&\textbf{Advice}&\textbf{Clarification}&\textbf{Multi-Turn}&\textbf{Expert Ann.}&\textbf{Source}\\
    \midrule
     CrimeKGAssistant&\checked &&  &&Real\\
 Hanfei&\checked & \checked &  \checked &&Gen.\\
 CAIL2023-ConvIR&& \checked &  \checked &&Gen.\\
 CAIL2024-ConGen&\checked & \checked &  \checked &&Gen.\\
\hline
     \textbf{LeCoDe (ours)}&\checked & \checked &  \checked  &\checked &Real\\
    \bottomrule
  \end{tabular}
  }
  \vspace{-2mm}
\end{table}

As shown in Table \ref{tbl:relatedwork}, our dataset (LeCoDe) distinguishes itself significantly from previous works.
LeCoDe uniquely combines lawyers' clarification questions and advice-giving capabilities while utilizing real-world scenarios, and the incorporation of expert annotation substantially enhances the dataset's value and quality.
Furthermore, we introduce a comprehensive evaluation framework that systematically evaluates LLMs' performance in legal consultation scenarios, incorporating two crucial interactive aspects: Clarification Capability and Advice Quality. 
These distinctive features collectively make LeCoDe a comprehensive and reliable resource for legal consultation tasks.

\section{LeCoDe}

\label{sec:lecode}
\subsection{Task Definition}
The legal consultation task is an interactive process between two agents: a \textbf{user client} $C$ and a \textbf{legal expert} $E$. 
Let $A_n = \{ a_1,...,a_n \}$ denote the set of atomic key facts privately held by $C$, $n$ is the cardinality of the set.
The client $C$ starts the consultation with a query $u_1^C$ about a specific case which may be vague and may not fully disclose all relevant facts in $A_n$.

The consultation dialogue sequence $D$ of length $T$ is defined as $D=\{(u_t^C,u_t^E)\}_{t=1}^T$, where $u_t^C$ and $u_t^E$ represent utterances of client $C$ and expert $E$ at turn $t$.
At each turn, the expert gives a response $u_t^E$ conditioned on: $u_t^E \sim p_E\{\cdot \mid D_{t-1},u_t^C\}$, where $D_{t-1}$ represents previous dialogue context. 
The expert's response type $r_t$ belongs to $R=\{ \text{Question}, \text{Advice} \}$, indicating whether to ask a clarifying question or provide legal advice.
Following the expert's query, the client's response is defined as $u_{t+1}^C \sim p_C\{\cdot \mid D_{t},A_n\}$, where the client generates responses based on the known key facts $A_n$ in response to the expert's latest clarifying question in dialogue context $D_t$.
The consultation terminates when either the expert provides legal advice or maximum turn is reached.

\textbf{Simulation Framework}:
To simulate real-world consultation interactions, the expert agent can be represented by various LLMs and strategies to generate clarifying questions or final advice based on the user's initial query and subsequent feedback.
To ensure reliable user simulation, we employ an LLM-based user agent that generates responses based on the complete key facts list $A_n$ and dialogue history.
The user agent follows several key principles: (1) providing concise responses to expert questions based on atomic key facts, (2) Reply ``unknown'' when uncertain, and (3) refusing to answer manipulative questions that attempt to elicit all known information.
To show the reliability of user agent, we validate various response LLMs and strategies on the performance of relevance and factuality (detailed settings and results in Appendix \ref{app:useragent}).

\subsection{LeCoDe Dataset Construction}
\label{subsec:lecodeconstruction}
The motivation of LeCoDe is to facilitate future research by establishing a high-quality and real-world legal consultation dialogue dataset.
Given the scarcity of publicly available real-time lawyer-client interaction data, we innovatively leverage Chinese short-video platforms, where licensed lawyers conduct live consultations with individuals seeking legal guidance, and subsequently publish representative cases as educational videos.

\begin{figure}[h]
    \centering
    \includegraphics[width=1.03\linewidth]{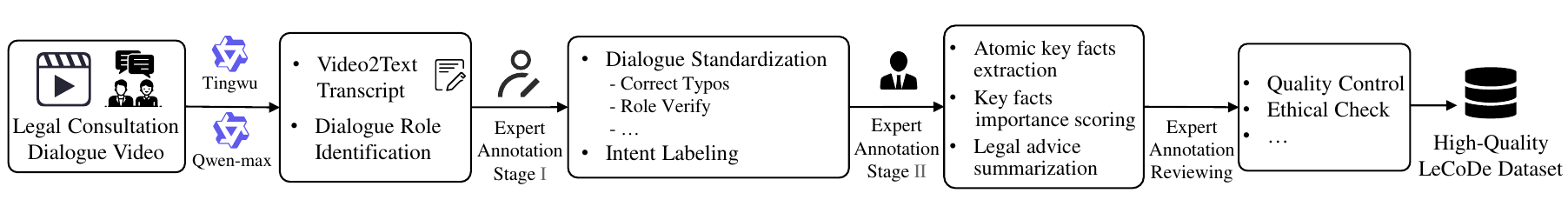}
    \caption{Data Construction Pipeline for LeCoDe.}
    \label{fig:annotationprocess}
\end{figure}

Our data consists of publicly available legal consultation videos from the Chinese short-video platforms. These short-videos are real-world live-stream consultations, with a total duration exceeding 235 hours.
To transform these video data into high-quality legal consultation dialogues, we design a systematic data processing pipeline, as illustrated in Figure \ref{fig:annotationprocess}.
The pipeline begins with processing using advanced LLMs - Tongyi Tingwu for speaker-aware transcription and Qwen-max for initial dialogue role identification.

\textbf{Two-stage expert annotation:} We then employ a two-stage expert annotation process to ensure data quality.
The first stage focuses on ensuring dialogue quality, where annotators perform dialogue standardization, transcription error correction, speaker role verification, and fine-grained utterance intent labeling. Among them, \textbf{intent labeling} annotation categorizes each utterance into one of ten predefined classes, such as ``Client Initial Query''  and ``Lawyer Clarifying Question.'' This process not only structures the dialogue data effectively but also improves the efficiency and accuracy of expert annotation in the subsequent stage.

The second annotation stage aims to enrich the dialogue dataset by leveraging legal expertise to create comprehensive training and evaluation resources.
Legal professionals serve as annotators to extract critical elements: clients' initial queries, atomic key facts $A_n$, the importance of key facts, and legal advice summarization. The detailed annotation content is as follows:

\textbf{Atomic key fact extraction} aims to identify the atomic factual units relevant to user needs from lengthy dialogues. On average, a single consultation contains around 9.19 key facts. In addition, the extracted atomic facts complement entity information and ensure consistency with the dialogue flow. During model training, these annotations assist in filtering out irrelevant content, enabling the model to learn effective and professional interaction strategies. In the evaluation phase, it serves as a means to assess the clarification capacity.

\textbf{Importance scoring of key facts} annotated the role of facts in legal analysis. The facts are categorized into three levels: Critical Facts (3 points), Secondary Facts (2 points), and Non-critical Facts (1 point). This scoring mechanism aims to help the model learn more efficient dialogue strategies and serves to evaluate whether it can focus on the core facts of a case.

\textbf{Legal advice summarization} aims to generate legal recommendations that are accurate, comprehensive, and concise. During training, this annotation enhances the model’s ability to produce fact-based, professional advice. In the evaluation phase, the task is used to assess the quality and effectiveness of the model’s generated advice.

Additionally, two expert annotation reviewers conducted thorough quality control and ethical compliance checks (see Section \ref{sec:ethicalconsederation}). To maintain rigorous annotation quality, we developed comprehensive annotation guidelines for both stages and conducted pilot annotations for annotator training.
Detailed annotation guidelines are provided in Appendix \ref{appendix:annotation}.
Specifically, we recruited 12 annotators with bachelor's degrees or above for the first stage, and 8 annotators with legal educational background for the second stage.
The inter-annotator agreement rates achieved during the pilot annotations were 88\% and 80\% for the first and second stages respectively, indicating substantial consistency in the annotation process.
For each dialogue, we paid annotators 8 CNY in stage-1 and 9 CNY in stage-2. The total annotation cost amounted to 63,360 CNY (\$8,751 at an exchange rate of 7.24 CNY/USD).

\subsection{LeCoDe Descrption}
In this section, we present comprehensive statistics of our constructed dataset, followed by an analysis of data distribution.

\textbf{Data Statistics:} LeCoDe consists of 3,696 dialogues split into training and test sets with an 8:2 ratio. Each dialogue contains an average of 29 turns and approximately 9 atomic key facts. The detailed data statistics are shown in Table \ref{tbl:dataset_distribution}, with illustrative examples provided in the Appendix \ref{appendix:subsection sampledata}.

\textbf{Data Distribution:} 
The dataset encompasses a diverse range of legal consultation categories, exhibiting the following distinctive characteristics:

\textit{\textbf{Diverse Legal Scenarios}}: As illustrated in Figure \ref{fig:data_distribution}d, LeCoDe covers three major legal domains: Criminal Law, Civil Law, and Administrative Law. Within Criminal Law, the most frequently consulted charges include Rape, Intentional Injury (In-Inj), Fraud, Intentional Homicide (In-Hom), and Theft. In Civil Law scenarios, the most common disputes involve Marriage (MRG), Contract (CNT), Personality Right (PRD), and Tort Liability (TLD).

\begin{figure}[htb]
  \begin{minipage}[b]{0.55\linewidth}
    \centering
    \includegraphics[width = \linewidth]{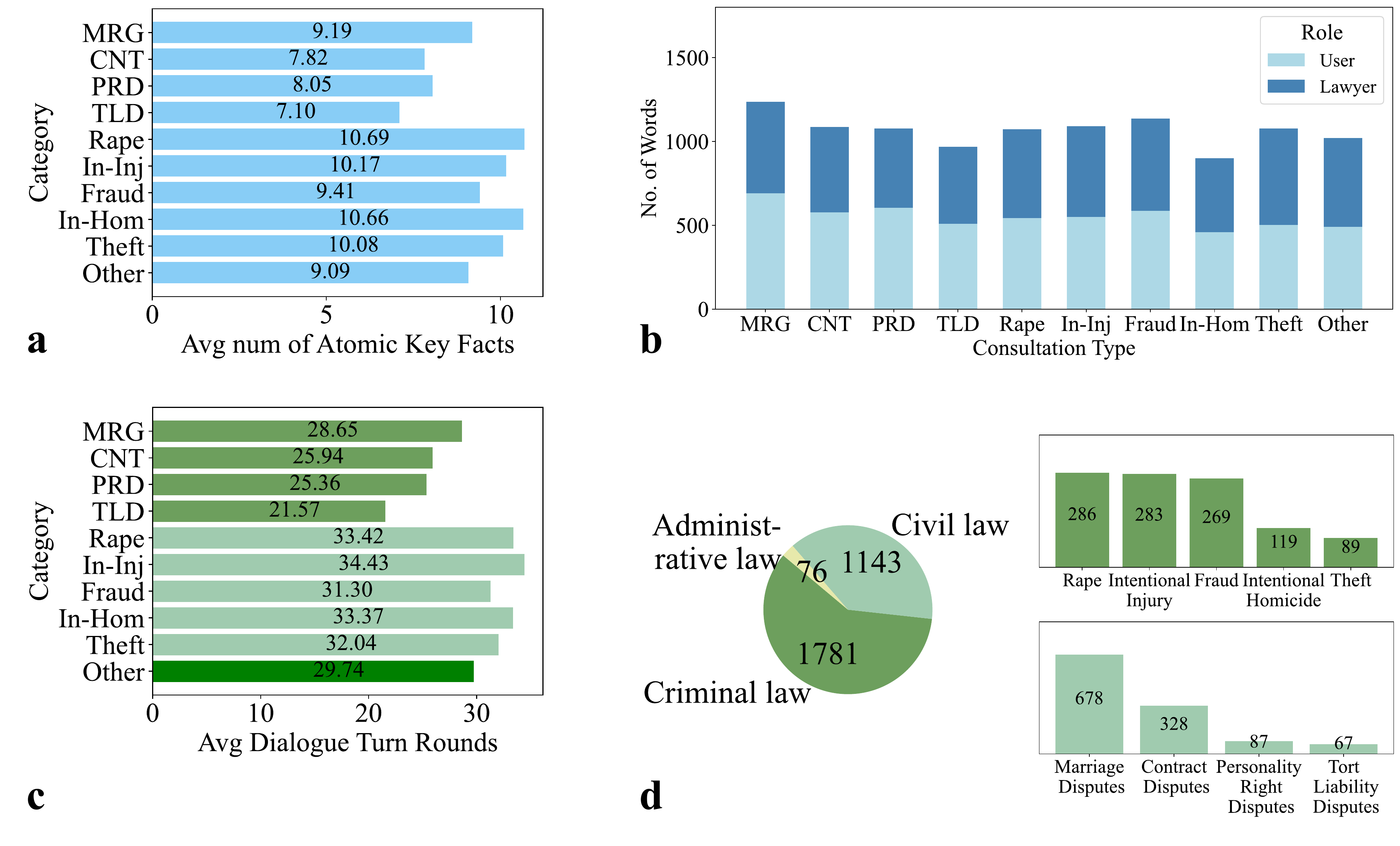}
    \caption{Dataset Distribution.}
    \label{fig:data_distribution}
  \end{minipage}\quad
  \begin{minipage}[b]{0.45\linewidth}
    \centering
        \captionof{table}{Statistics over LeCoDe Dataset.}
    \scalebox{0.7}{
    \begin{tabular}{llll}
      \toprule
        Dataset     & LeCoDe     & Train Set & Test Set \\
        \midrule
        \# Dialogue Samples & 3,696  & 2,956& 740     \\
        \# Total turns     & 110,008 & 88,342& 21,666      \\
         avg turns per dialogue     &29.76       & 29.89& 29.28  \\
         avg key facts per dialogue  &9.19      & 9.24& 9.01 \\
        \bottomrule
        &&&\\
         &&&\\
         &&&\\
         &&&\\
         &&&\\
    \end{tabular}}
    \label{tbl:dataset_distribution}
  \end{minipage}
\end{figure}

\textit{\textbf{Varying Complexity Across Domains}}: Figure \ref{fig:data_distribution}a presents the average number of key facts per dialogue across common consultation scenarios.
Notably, criminal law cases contain more facts compared to civil law cases, reflecting the inherent complexity of criminal proceedings.
This pattern is also reflected in Figure \ref{fig:data_distribution}c, where criminal law consultations average 23.7 turns compared to 19.6 turns in civil law cases.

\textit{\textbf{Balanced Dialogue Dynamics}}: Figure \ref{fig:data_distribution}b presents the word count distribution between participants. Interestingly, users' contributions are comparable to, or slightly exceed, those of lawyers.
This pattern reflects the realistic nature of legal consultations. Clients often have limited legal knowledge and are unable to present comprehensive, well-structured requirements in the initial query. Instead, they gradually articulate their situations in response to lawyers’ clarifying questions.

Overall, LeCoDe authentically reflects real-world legal consultations through its diverse case categories and natural dialogue patterns.

\subsection{Evaluation Metrics}
\label{subsec:evaluationmetrics}

We evaluate expert responses from two main perspectives. 
First, we assess \textbf{Clarification Capability} through \textit{effectiveness} and \textit{efficiency} metrics.
Second, we evaluate \textbf{Advice Quality} using both \textit{automated} and \textit{LLM-based evaluation} metrics.

\textbf{Clarification Capability:}
\textit{Effectiveness:} we measure the model's ability to elicit ground truth atomic key facts ($A_n$) through clarification questions during dialogue simulation.
To evaluate the coverage of key facts through expert questioning, we employ multiple metrics: \textbf{Recall (R)} assesses whether all key facts are acquired under the guidance of the model, while \textbf{Weighted Recall (WR)} considers the importance score of each key fact.
Given that high-quality legal consultation requires uncovering key information in the early rounds, we utilize \textbf{Recall@5 (R@5)} to evaluate how effectively the first five questions elicit key facts, and \textbf{NDCG} to assess whether the model asks questions in order of fact importance.
\textit{Efficiency:} we measure the average number of dialogue turns (\textbf{AT}).

\textbf{Advice Quality:}
we evaluate the semantic alignment between the generated and reference advice using both \textit{automated metrics} \textbf{ROUGE-L} (R-L)~\citep{lin-2004-rouge} and \textbf{BERTScore} (BS)~\citep{bertscore} and \textit{LLM-based evaluation} metrics. The latter assesses five aspects: \textbf{Professionalism (Pro)}, \textbf{Fluency (Flu)}, \textbf{Completeness (Com)}, \textbf{Satisfaction (Sat)}, and \textbf{Safety (Safe)}, culminating in a \textbf{Overall Score (OA)}.

Detailed definitions and calculation methods for all metrics are provided in the Appendix \ref{app:evaluation metrics}.

\subsection{Training Dialogue Construction Strategy for SFT}
\label{section:SFT-strategy}

\begin{wrapfigure}{r}{0.6\textwidth}
    \includegraphics[width=1\linewidth]{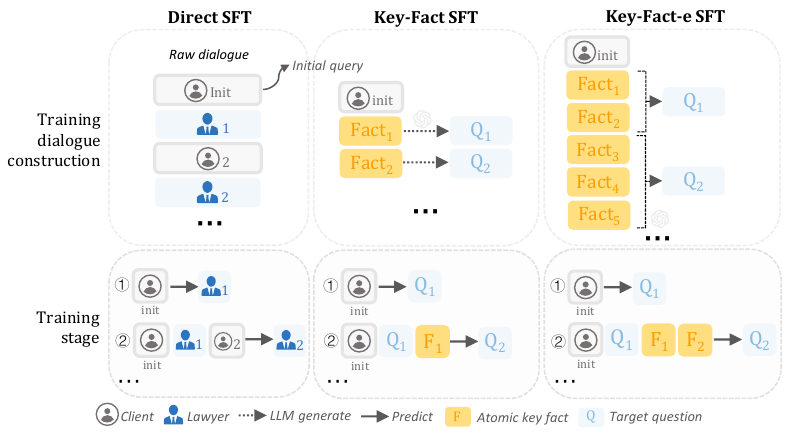}
    \caption{Training Dialogue Construction Strategy for SFT}
    \label{fig:sft_strategy}
\end{wrapfigure}

To fully utilize LeCoDe training data, we propose three strategies to construct training multi-turn dialogue for fine-tuning LLMs, as shown in Figure \ref{fig:sft_strategy}.

\textbf{Direct SFT} directly leverages LeCoDe training data to predict lawyer utterances based on dialogue history, simulating natural consultation flow.

\textbf{Key-fact SFT} employs a targeted approach where stronger LLMs first generate multiple target questions for each atomic key fact. The target model is then trained to learn how to generate these target questions, establishing a one-to-one mapping between questions and facts.

\textbf{Key-fact-e SFT}, an enhanced version of the previous approach, also utilizes stronger LLMs to generate target questions that can effectively cover multiple (1-3) key facts simultaneously. Training data is then constructed based on these questions and corresponding key facts, enabling more efficient information elicitation.
The detailed implementations of the strategies are provided in the Appendix \ref{app:sftstrategy}.

\subsection{Legal and Ethical Consideration}
\label{sec:ethicalconsederation}
Considering the sensitivity of legal domain, we adopted strict measures to ensure ethical compliance and minimize potential risks.
First, our dataset exclusively consists of publicly shared educational videos on the Chinese short-video platforms, where all videos have undergone strict platform moderation and ethical review, ensuring both anonymity and value alignment.
Second, legal experts conducted thorough content reviews to filter out potentially discriminatory, violent, or offensive content, while ensuring anonymity and mitigating potential biases.
Furthermore, we release LeCoDe under a license restricted to academic research purposes only (see in Appendix \ref{app:license}).
We believe these measures effectively mitigate potential social risks.
A detailed discussion of potential impacts is provided in Appendix \ref{app:discussion}.

\section{Experiment}

\subsection{Baselines}
We evaluate models across three distinct categories: (1) \textbf{Closed-source Commercial LLMs}, including GPT series models (GPT-4~\citep{achiam2023gpt}, GPT-4o~\citep{hurst2024gpt4o}, GPT-3.5-turbo) and Qwen series models (Qwen-max~\citep{qwen2.5}, Qwen-turbo~\citep{qwen2.5}); (2) \textbf{Open-source LLMs}, comprising DeepSeek-R1~\citep{deepseekai2025deepseekr1incentivizingreasoningcapability}, DeepSeek-V3~\citep{deepseekai2024deepseekv3technicalreport}, Qwen2.5-72B~\citep{qwen2.5}, Qwen2.5-7B~\citep{qwen2.5}, Llama-3.1-8B~\citep{grattafiori2024llama3}, and GLM-4-9B~\citep{glm2024chatglm4}, GLM-4-32B~\citep{glm2024chatglm4}; and (3) \textbf{Legal-domain Specialized Models and Strategies}, which include domain-specific LLMs (ChatLaw-13B~\citep{cui2023chatlaw}, Lawyer-LLaMA~\citep{huang2023lawyer}, Tongyi-Farui~\citep{Tongyi-farui}). 
The clarifying strategy incorporates \textbf{MediQ}~\citep{li2024mediq}, a question-asking framework adapted from clinical reasoning to legal consultation. 
The detailed setting can refer to Appendix \ref{app:experiment setting}.
Additionally, we propose three training dialogue construction strategies for supervised fine-tuning (SFT) to enhance LLMs' legal consultation capabilities:
\textbf{Direct SFT}, \textbf{Key-fact SFT} and \textbf{Key-fact-e SFT}, as described in Section \ref{section:SFT-strategy}.

\subsection{Experiment Setting}
\label{experiment setting}

For user simulation, LLM-based agents have demonstrated strong response capabilities~\citep{usersimulator2,li2024mediq,usersimulator-2024-reliable}.
In our setting, an ideal user simulator should exhibit both relevance (responding appropriately to lawyers' questions) and factuality (adhering to given key fact without hallucination).
After experimenting with various response strategies and LLMs (detailed settings and results in Appendix \ref{app:useragent}), we use Qwen-max with First-Person Given Dialogue Instruction strategy for user simulation, which achieves best performance on relevance and factuality.

For expert agents, we implement both zero-shot and few-shot strategies across all general LLMs.
In few-shot scenarios, we manually crafted two demonstrations.
The maximum turn is 10 rounds.
If the maximum turn limit is reached, the LLMs will be instructed to generate final legal advice.
All prompt templates and settings are provided in the Appendix \ref{app:lawyeragent}.
We use the evaluation metrics in Section \ref{subsec:evaluationmetrics} for evaluation, and utilize GPT-4o-mini for advice quality assessment (refer to Appendix \ref{app:prompt_template}).

\subsection{Experiment Results}

We report zero-shot performance of all models in Table \ref{tbl:zs_mainresult1}, with few-shot results provided the Tables \ref{tbl:app_fs_mainresult1}-
\ref{tbl:app_fs_mainresult2} in Appendix. Our experiments reveal several key findings:

\textbf{Clarification Capability:}
\textbf{(1) }Closed-source commercial LLMs generally outperform open-source LLMs.
GPT-4 achieves the best balance between efficiency and effectiveness, demonstrating strong performance in questioning ability.
\textbf{(2) }Qwen-max and Qwen2.5-72B show notable performance in Recall and Weighted Recall.
\textbf{(3) }While MediQ effectively reduces dialogue turns, it compromises question quality. Domain-specific models like ChatLaw and Lawyer-LLama often bypass the questioning phase and provide direct advice due to poor instruction following, resulting in shorter but less effective consultations.
\textbf{(4) }Notably, SFT strategies significantly outperform other approaches across all clarification metrics, achieving remarkable Recall (53.8\%) and NDCG (84.8\%) scores.

\begin{table*}[ht]
  \caption{Main results on zero-shot setting, where top-6 scores are marked in \colorbox{lightblue!95}{blue}, the highest is \textbf{bolded} and the second-highest is \underline{underlined}.}
  \label{tbl:zs_mainresult1}
  \centering
  \scalebox{0.77}{
  \begin{tabular}{l|cccc|c|cc|cccccc}
    \toprule
    \multirow{3}{*}{\textbf{Models}}& \multicolumn{5}{c|}{\textbf{Clarification Capability}}& \multicolumn{8}{c}{\textbf{Advice Quality}}\\
    \cline{2-14}
 & \multicolumn{4}{c|}{\textbf{Effe.}}& \textbf{Effi.}  & \multicolumn{2}{c|}{\textbf{Auto.}}& \multicolumn{6}{c}{\textbf{LLM-Eval.}}\\
 & R& WR& R@5& NDCG& AT & R-L& BS& Pro& Flu& Com& Sat& Safe&OA\\
    \midrule
     \multicolumn{14}{c}{\textit{Closed-source LLMs}}\\
     \hline
 GPT-4& \colorbox{lightblue!95}{35.9}& \colorbox{lightblue!95}{37.2}& \colorbox{lightblue!95}{\underline{34.4}}& \colorbox{lightblue!95}{\underline{74.7}}&\colorbox{lightblue!95}{4.9} & \colorbox{lightblue!95}{\underline{17.5}}& \colorbox{lightblue!95}{\underline{65.9}}& \colorbox{lightblue!95}{56.8}& \colorbox{lightblue!95}{71.9}& \colorbox{lightblue!95}{49.9}& \colorbox{lightblue!95}{55.8}& \colorbox{lightblue!95}{71.1}&\colorbox{lightblue!95}{58.1}\\
 GPT-4o& 33.0& 33.9& 25.9& 63.2& 10 & 13.0& 62.9& 50.1& 69.6& 44.7& 52.4& 66.6&53.5\\
 GPT-3.5& 26.9& 28.0& 23.6& 67.1& 9.2 & 12.4& 63.1& 51.8& 67.6& 45.8& 51.6& 66.0&53.8\\
 Qwen-max& \colorbox{lightblue!95}{\underline{39.8}}& \colorbox{lightblue!95}{\underline{41.3}}& \colorbox{lightblue!95}{31.2}& \colorbox{lightblue!95}{70.7}& 10 & 13.9& 63.9& 54.0& 71.4& 47.6& \colorbox{lightblue!95}{55.3}& 68.9&56.2\\
 Qwen-turbo& 27.3& 28.3& 23.2& 61.2& 9.4 & 10.6& 61.8& 44.8& 62.7& 40.0& 47.6& 61.6&48.5\\
 \hline
 \multicolumn{14}{c}{\textit{Open-source LLMs}}\\
     \hline
    deepseek-v3& 28.0&29.0& 25.1& 65.4&  7.5 & \colorbox{lightblue!95}{16.2}& \colorbox{lightblue!95}{65.3}& \colorbox{lightblue!95}{57.6}& \colorbox{lightblue!95}{72.9}& \colorbox{lightblue!95}{50.8}& \colorbox{lightblue!95}{\underline{56.2}}& \colorbox{lightblue!95}{72.1}&\colorbox{lightblue!95}{58.9}\\
    deepseek-r1& 27.6&28.8& 25.1& 63.9&  \colorbox{lightblue!95}{6.3} & 12.4& 64.0& \colorbox{lightblue!95}{\textbf{63.8}}& \colorbox{lightblue!95}{\underline{74.1}}& \colorbox{lightblue!95}{\textbf{55.5}}& \colorbox{lightblue!95}{55.9}& \colorbox{lightblue!95}{\textbf{76.5}}&\colorbox{lightblue!95}{\textbf{62.2}}\\
 Qwen2.5-72B& \colorbox{lightblue!95}{39.6}& \colorbox{lightblue!95}{40.8}& \colorbox{lightblue!95}{30.4}& \colorbox{lightblue!95}{71.9}& 10 & 14.7& 63.8& 54& \colorbox{lightblue!95}{71.6}& 48.5& 55.2& 69.4&56.6\\
 Qwen2.5-7B& 31.1& 32.1& 27.7& \colorbox{lightblue!95}{70.1}& 10 & 9.2& 60.0& 42.3& 60.4& 37.6& 45.0& 57.8&45.7\\
 Llama-3.1-8B& 22.1& 23.1& 19.1& 55.0& 7.8 & 13.9& 63.1& 51.7& 69.1& 46.5& 51.7& 67.1&54.1\\
 GLM4-32B& 34.1& 35.4& \colorbox{lightblue!95}{28.8}& 67.3& 7.3 & 14.3& 64.1& \colorbox{lightblue!95}{55.6}& 70.4& \colorbox{lightblue!95}{49.0}& 54.3& \colorbox{lightblue!95}{70.4}&\colorbox{lightblue!95}{56.7}\\
 GLM4-9B& 34.2& 35.3& 26.1& 65.0& 10 & 13.9& 63.1& 51.7& 69.1& 46.5& 51.7& 67.1&54.1\\
 \hline
  \multicolumn{14}{c}{\textit{Domain-specific LLMs or Clarifying Strategy}}\\
     \hline
 ChatLaw& 16.7& 17.4& 16.3& 47.8& \colorbox{lightblue!95}{4.4} & 7.8& 53.0& 28.8& 40.0& 25.5& 31.0& 39.4&30.7\\
 Lawyer-LLaMA& 21.8& 22.9& 21.8& 62.9& \colorbox{lightblue!95}{\underline{4.2}} & 12.4& 59.7& 37.6& 52.7& 33.1& 39.0& 51.8&39.4\\
  Farui& 28.4& 29.5& 25.1& 64.2& 8.9 & \colorbox{lightblue!95}{16.6}& \colorbox{lightblue!95}{64.5}& 52.2& 66& 47.4& 51.3& 66.3&53.8\\
 MediQ& 22.1& 22.9& 19& 53.5& \colorbox{lightblue!95}{\textbf{4.2}} & 13.3& 63.7& 51.8& 69.1& 45.5& 51.9& 66.7&54.0\\
 Direct SFT& \colorbox{lightblue!95}{37.2}& \colorbox{lightblue!95}{37.6}& 25.6& 62.3& 9.6 & \colorbox{lightblue!95}{15.4}& \colorbox{lightblue!95}{64.9}& 51.7& 65.3& 45.6& 50.1& 63.1&52.4\\
     Key-fact SFT&\colorbox{lightblue!95}{\textbf{53.8}}&\colorbox{lightblue!95}{\textbf{55.1}}& \colorbox{lightblue!95}{43.5}& \colorbox{lightblue!95}{84.7}&  9.2 & \colorbox{lightblue!95}{18.5}& \colorbox{lightblue!95}{67.5}& \colorbox{lightblue!95}{59.3}& \colorbox{lightblue!95}{73.0}& \colorbox{lightblue!95}{\underline{52.3}}& \colorbox{lightblue!95}{55.9}& \colorbox{lightblue!95}{71.2}&\colorbox{lightblue!95}{59.4}\\
    Key-fact-e SFT&\colorbox{lightblue!95}{51.0}&\colorbox{lightblue!95}{51.1}& \colorbox{lightblue!95}{\textbf{45.3}}& \colorbox{lightblue!95}{\textbf{84.8}}&   \colorbox{lightblue!95}{6.6} & \colorbox{lightblue!95}{\textbf{19.1}}& \colorbox{lightblue!95}{\textbf{67.6}}& \colorbox{lightblue!95}{\underline{59.3}}& \colorbox{lightblue!95}{\textbf{73.6}}& \colorbox{lightblue!95}{52.0}& \colorbox{lightblue!95}{\textbf{56.7}}& \colorbox{lightblue!95}{\underline{72.1}}&\colorbox{lightblue!95}{\underline{59.5}}\\
    \bottomrule
  \end{tabular}
}
\end{table*}

\textbf{Advice Quality:}
\textbf{(1) }GPT-4 leads among closed-source LLMs across multiple advice quality metrics. 
\textbf{(2) }Deepseek-R1 excels among open-source LLMs, even achieves SOTA performance on LLM-eval metrics.
\textbf{(3)} SFT strategies demonstrate superior performance in both advice quality metrics.

Our analysis in Tables \ref{tbl:app_fs_mainresult1}-
\ref{tbl:app_fs_mainresult2} show that few-shot prompting negatively impacts clarification capability in most models (6/11), while improving advice quality (8/11).
This suggests that while few-shot examples may constrain questioning scope, they provide useful demonstrations for advice generation.
Overall, \textbf{LLMs show suboptimal performance}. While \textbf{SFT strategies show effectiveness, there remains substantial room for improvement}, particularly in clarification capability (with only 53.8\% recall) and the quality of advice.

\subsection{Further Analysis}
In this section, we analyze key findings from our experiments to derive insights for future studies in legal consultation scenarios and potential improvements.

\begin{figure}[htbp]
    \centering
    \begin{minipage}{0.96\linewidth}  
        \centering
        \begin{minipage}{0.63\linewidth}
            \centering
            \includegraphics[width=\linewidth]{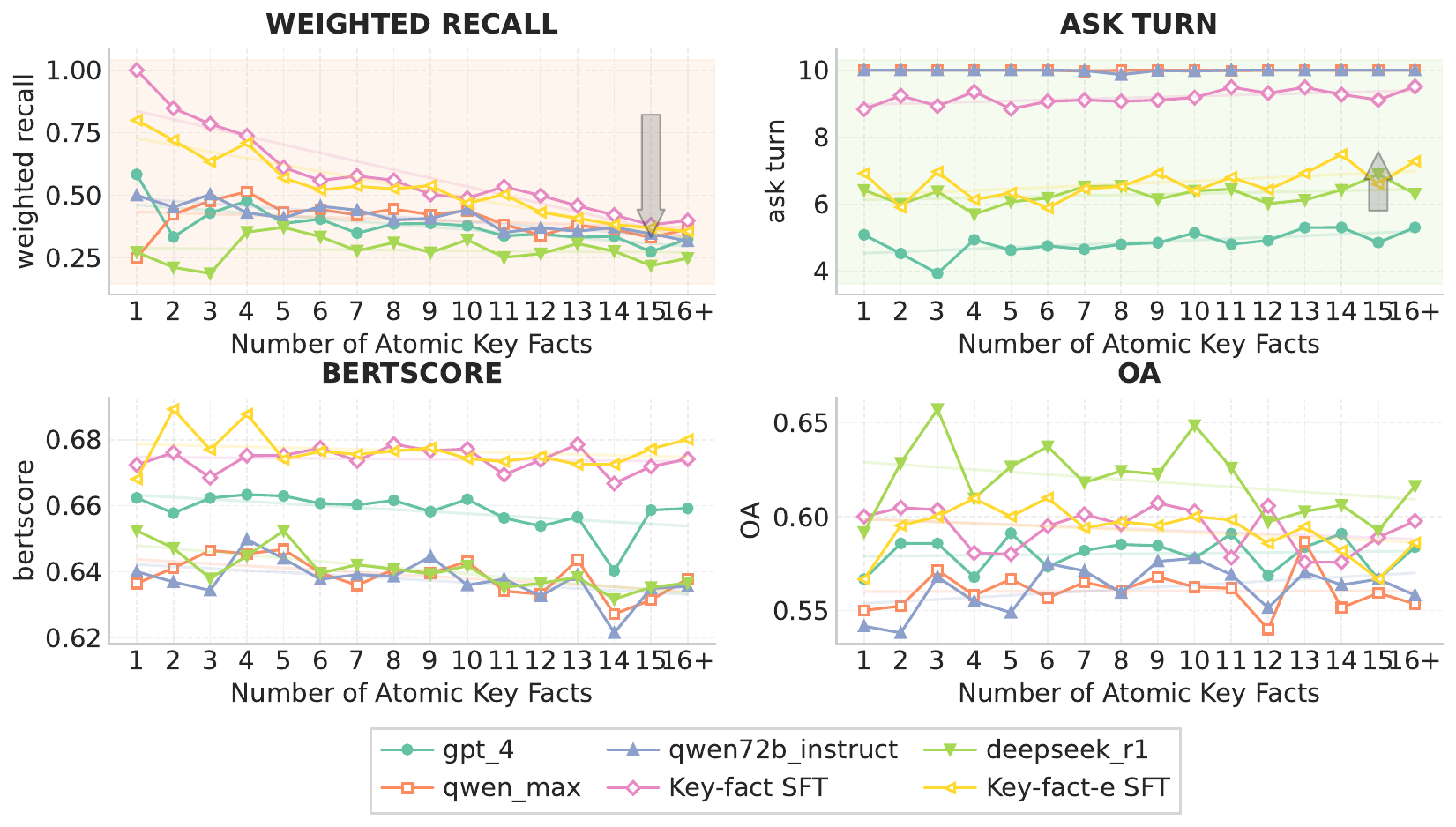}
            \caption{Impact of Case Complexity on Model Performance Metrics: Analysis of Weighted Recall, Ask Turn, BERTScore, and Overall Score (OA) across Different Numbers of Atomic Key Facts.}
            \label{fig:performance-label}
        \end{minipage}
        \hfill
        \begin{minipage}{0.33\linewidth}
            \centering
            \includegraphics[width=\linewidth]{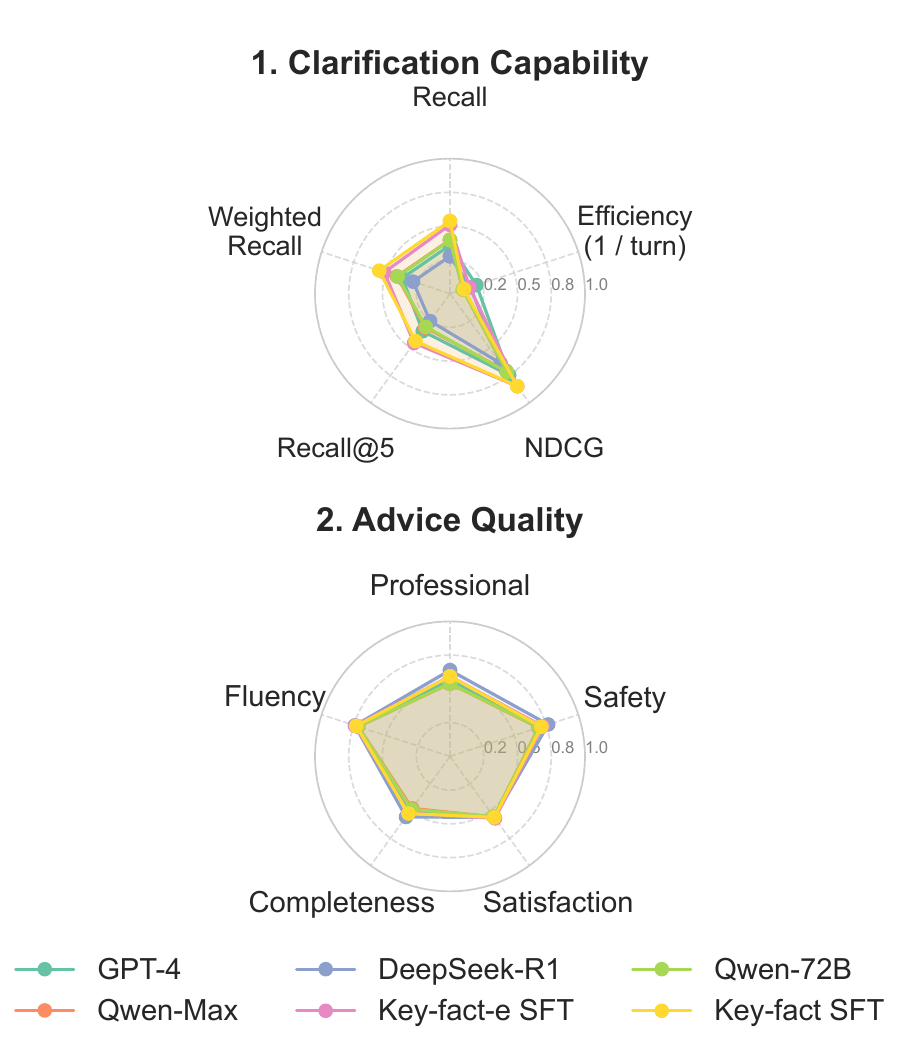}
            \caption{Radar chart showing the LLMs performance.}
            \label{fig:radar-label}
        \end{minipage}
    \end{minipage}
    \vspace{-5mm}
\end{figure}

As illustrated in Figure \ref{fig:performance-label}, we analyze the relationship between case complexity (measured by the number of atomic key facts) and model performance, including Weighted Recall (WR) for clarification effectiveness, Ask Turn for efficiency, and BERTScore (BS) and Overall Accuracy (OA) for advice quality. 
The results reveal several notable patterns: As case complexity increases, a significant decline in WR and subtle decreases in both BS and OA, \textbf{indicating that performance declines with case complexity}.
Interestingly, the Ask Turn metric shows an upward trend, \textbf{suggesting that LLMs exhibit some awareness of case complexity}.
These findings suggest LLMs can be improved to better assess case complexity and ask more effective questions in the future.

\textbf{Significant room for improvement in LLMs' legal consultation capabilities}. As shown in Figure \ref{fig:radar-label}, nearly all LLMs achieve only approximately 30\%-50\% recall in clarification capability. Moreover, substantial gaps remain in advice quality, including professional, user satisfaction, and completeness.

\begin{wrapfigure}{r}{0.57\textwidth}
    \includegraphics[width=1\linewidth]{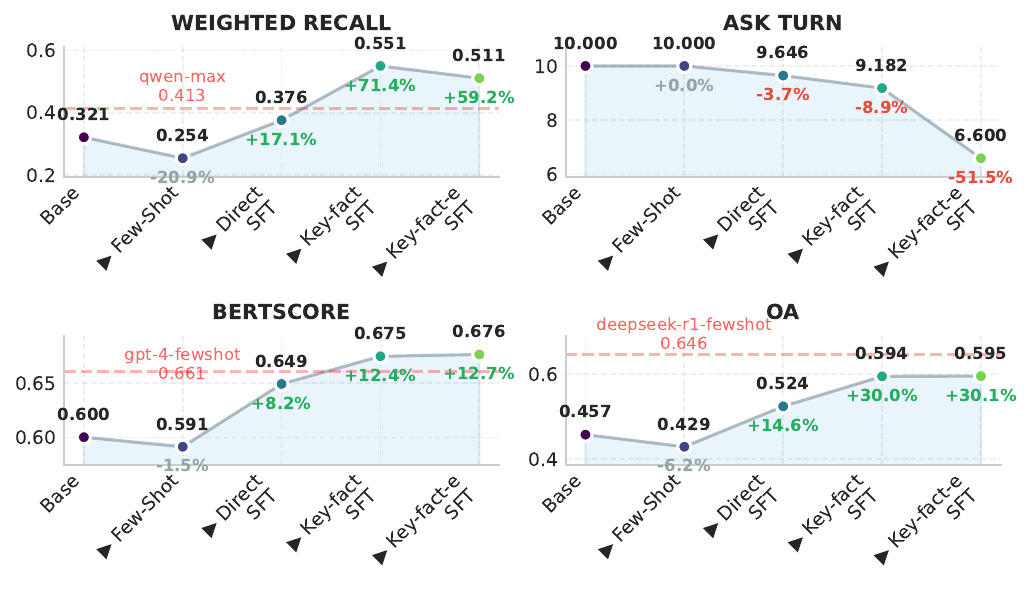}
    \caption{Comparison of Different Strategies.}
    \label{fig:evolution}
\end{wrapfigure}

\textbf{How to effectively enhance LLMs' legal consultation abilities?}
As shown in Figure \ref{fig:evolution}, using Qwen2.5-7B as our base model, it reflects few-shot strategies consistently produced adverse effects, suggesting that conventional \textbf{prompt engineering approaches may be insufficient for meaningful improvements}.
Consequently, we propose SFT strategies to rapidly and effectively develop consultation capabilities from the dataset. 
The results show that Key-fact SFT, which trains LLMs to target specific key facts through strategic questioning, significantly outperforms Direct SFT's raw dialogue pattern learning approach. Notably, Key-fact-e, achieves remarkable improvements across all metrics (WR: +59.2\%, BS: +12.7\%, OA: +30.1\%) while reducing dialogue turns by 51.5\%, even surpassing the performance of SOTA LLMs.

\section{Conclusion}

We introduce LeCoDe, the first dataset of real-world legal consultation dialogues, with an evaluation framework measuring LLMs' clarification capability and advice quality. Our analysis reveals significant limitations of existing LLMs in legal consultation tasks. Further, we propose supervised fine-tuning strategies that lead to notable improvements in the legal consultation performance of LLMs.
In the future, we encourage researchers to explore two promising avenues: (1) exploring more strategies to enhance LLMs' consultation capabilities and (2) effectively incorporating external legal knowledge. Through coordinated advancement, these efforts will expand the potential of professional services to become more accessible and effective.

\clearpage
\newpage
\bibliographystyle{plain}
\bibliography{custom.bib}



\clearpage
\appendix
\newpage

\appendix
\addcontentsline{toc}{section}{Appendices}  
\startcontents[appendix]
\begin{center}
    \large\bfseries Appendix Contents  
\end{center}
\vspace{10pt}  
\printcontents[appendix]{}{1}{}  

\section{Resource Availability and License}
\label{app:availablility}
\subsection{Resource Availability}
Upon acceptance, we will make the data and source code publicly available.

\subsection{Licenses}
\label{app:license}
This work is licensed under a Creative Commons Attribution- NonCommercial-NoDerivatives 4.0 International License (CC BY-NC-ND 4.0).
\textbf{All resources are for scientific research only.}

\section{Discussion}
\label{app:discussion}
\subsection{Limitations}

Our work has several limitations and opportunities for future research. 
First, the current dataset primarily consists of Chinese legal consultation dialogues. We hope to incorporate legal consultation data from diverse jurisdictions and legal systems to evaluate LLMs' capabilities across different legal contexts in the future.

Second, while our framework focuses on professional legal consultation capabilities - specifically clarification capability and professional advice quality - real-world consultations are often more complex and require chit-chat and emotional support, aspects that could be incorporated into future evaluation frameworks.

Third, although we focus on legal consultations, similar interactive consultation scenarios exist in other professional domains, such as healthcare.
We plan to validate our framework extension beyond legal applications.

Finally, while our SFT-based strategies show improvements over baseline LLM performance, the overall results remain suboptimal. 
We encourage researchers to explore additional approaches for enhancing model capabilities, such as reinforcement learning techniques to optimize consultation strategies, external knowledge integration through RAG or knowledge graph approaches, and other methods for improving model performance in professional consultation scenarios.

\subsection{Broader Impact}
\label{app:broaderimpact}

We propose LeCoDe to help practitioners better understand and evaluate LLMs' performance in legal consultation scenarios, providing foundational data and evaluation framework for future research.
However, given the sensitive and high-risk nature of the legal domain, we must carefully consider potential risks. While we actively explore LLMs' applications in legal scenarios, we must guard against potential unfairness and societal disruption. The dataset is intended purely for academic research, not as a replacement for expert legal consultation. Legal advisory requires rich professional knowledge, practical experience, and human insight that no LLM or AI technology can substitute.

LeCoDe comprises real-world consultation data, and we have made extensive efforts to filter out potentially discriminatory, violent, or offensive content while ensuring anonymity and mitigating potential biases. However, we acknowledge that using LeCoDe for model training may still introduce potential biases and discrimination. Therefore, we emphasize that the dataset is licensed exclusively for academic research and prohibited from commercial or practical applications.

To mitigate potential negative impacts, we explicitly state the dataset's usage restrictions and limitations, and continuously monitor the practical applications of research outcomes to adjust recommendations accordingly. 
Ultimately, we hope this work contributes to the intelligent development of legal services while ensuring technological advancement serves the goals of social fairness and justice.

\section{User Agent}
\label{app:useragent}
To ensure our LLM-based agent for user simulation, we propose several strategies and evaluate the reliability for user agent.
We first introduce the evaluation metrics, then evaluation settings and corresponding prompt template, finally show the evaluation results for user agent reliability.

\subsection{Evaluation Metrics for User Agent}
While LLM-based models have demonstrated strong response capabilities~\citep{li2024mediq}, our evaluation focuses on two critical aspects of an ideal user simulator: \textbf{relevance} (appropriate responses to lawyers' questions) and \textbf{factuality} (adherence to given information without hallucination).

\textbf{Relevance} measures the user agent's ability to comprehend and provide pertinent responses to lawyers' clarifying questions, evaluating the model's question understanding capabilities.

\textbf{Factuality} assesses whether the user's responses to clarifying questions align with the provided atomic fact background information, measuring the model's ability to accurately extract and utilize known information while avoiding fabrication.

\subsection{Evaluation Setting for User Agent}

We randomly selected 100 initial queries with their corresponding 5-round dialogue contexts from the training set, for which we manually generated three clarifying questions. The evaluation was conducted using various Qwen-series models, including Qwen2.5-7B, Qwen2.5-32B, Qwen2.5-72B, and Qwen-max. GPT-4o-mini served as the judge for assessing both relevance and factuality.

We evaluated three prompt engineering strategies:
\begin{itemize}
\item \textbf{Instruct Given Dialogue}: Providing complete interaction dialogue history and atomic key facts to instruct LLM responses
\item \textbf{Instruct Given Question}: Providing only the current question and atomic key facts for LLM responses
\item \textbf{First-person Given Dialogue}: Instructing LLMs to simulate first-person responses based on complete interaction dialogue
\end{itemize}

The prompt template for each strategies can be seen in Appendix \ref{app:user_prompt_template_for_each_PE}.
The evaluation prompt template for GPT-4o-mini can be seen in Appendix \ref{app:Prompt_Templates_for_LLM-as-a-judge_for assessing_relevance_and_factuality}.

\subsection{Prompt Templates for each PE Strategies}
\label{app:user_prompt_template_for_each_PE}
\begin{tcolorbox}[title={Instruct Given Dialogue (translated from Chinese)}, colframe=gray]
\small\texttt{\textcolor{teal}{You are a honest and reliable legal assistant who understands user-related information and attempts to answer lawyers' questions about users.}\vspace{4mm}\\
The following is [Atomic Key Facts] describing user information\\
\{Atomic Key Facts\}\\ \\
The following is the interactive dialogue between you (user) and the lawyer\\
\{current dialogue\}\\ \\
- Use the above key fact to answer the lawyer's questions, selecting no more than 3 answers.\\
- If the above key fact cannot answer the lawyer's questions, [only reply "I Don't know"].\\ 
- Only answer what is asked in the question, without providing any analysis, speculation or conclusions. \\
- Only select all content from the above information that can answer the question, and only that.
}\normalsize
\end{tcolorbox}

\begin{tcolorbox}[title={Instruct Given Question (translated from Chinese)}, colframe=gray]
\small\texttt{\textcolor{teal}{You are a honest and reliable legal assistant who understands user-related information and attempts to answer lawyers' questions about users.}\vspace{4mm}\\
The following is [Atomic Key Facts] describing user information\\
\{Atomic Key Facts\}\\ \\
The following is the lawyer's question you need to answer\\
\{lawyer question\}\\ \\
- Use the above key fact to answer the lawyer's questions, selecting no more than 3 answers.\\
- If the above key fact cannot answer the lawyer's questions, [only reply "I Don't know"].\\ 
- Only answer what is asked in the question, without providing any analysis, speculation or conclusions. \\
- Only select all content from the above information that can answer the question, and only that.
}\normalsize
\end{tcolorbox}

\begin{tcolorbox}[title={First-person Given Dialogue (translated from Chinese)}, colframe=gray]
\small\texttt{\textcolor{teal}{You are simulating a real user in a legal consultation scenario.\\
Your role is a consultant [user] who encounters legal difficulties and seeks legal help. Since you lack sufficient legal knowledge, you cannot raise a clear, professional legal question. Therefore, you need to respond to the lawyer's questions based on the [Atomic Key Facts] you know.
}\vspace{4mm}\\
The following is [Atomic Key Facts] describing user information\\
\{Atomic Key Facts\}\\ \\
The following is the interactive dialogue between you (user) and the lawyer\\
\{current dialogue\}\\ \\
Your task is to answer relevant questions based on the [Atomic Key Facts] description.\\
Note\\
<1>Your role is set as seeking legal help, and you cannot use professional legal terminology.\\
<2>Only answer the lawyer's questions based on the content of [Atomic Key Facts], do not generate other content. Select no more than 3 key pieces of information to answer in one reply.
If [Atomic Key Facts] cannot answer the question, then [only reply "I Don't know"].\\
<3>For any question you're unsure about, [only reply "I Don't know"].\\
<4>Keep replies as brief as possible.\\
<5>Prevent excessive information disclosure.
}\normalsize
\end{tcolorbox}

\subsection{Prompt Templates for LLM-as-a-judge for assessing relevance and factuality}
\label{app:Prompt_Templates_for_LLM-as-a-judge_for assessing_relevance_and_factuality}
\begin{tcolorbox}[title={LLM-as-a-judge for assessing relevance (translated from Chinese)}, colframe=gray]
\small\texttt{\textcolor{teal}{You will receive a [Clarifying Question] raised by a lawyer in the dialogue, and a [User Response] based on [Atomic Facts] background information. Your task is to perform a binary classification (0 or 1) on the user's response, judging its relevance to the lawyer's [Clarifying Question].}\vspace{4mm}\\
In some cases, the user may reply "I Don't know" based on the lawyer's advice.\\
If the user replies "I Don't know", you need to determine whether the [Clarifying Question] cannot be answered based on the content of [Atomic Key Facts]. If it indeed cannot be answered, then the relevance of replying "Don't know" should be 1; otherwise, it's 0.\\
\text{[Dialogue Context]}\\
\{dialogue\}\\ \\
\text{[Atomic Key Facts]}\\
\{Atomic Key Facts\}\\ \\
\text{[Clarifying Question]}\\
\{lawyer question\}\\ \\
\text{[User Response]}\\
\{User Response\}\\ \\
}\normalsize
\end{tcolorbox}

\begin{tcolorbox}[title={LLM-as-a-judge for assessing factuality (translated from Chinese)}, colframe=gray]
\small\texttt{\textcolor{teal}{You will receive a [User Response] based on [Atomic Key Facts] background information. Your task is to perform a binary classification (0 or 1) on the factuality of the user's response, determining whether it is consistent with the provided atomic facts background information.}\vspace{4mm}\\
\text{[Dialogue Context]}\\
\{dialogue\}\\ \\
\text{[Atomic Key Facts]}\\
\{Atomic Key Facts\}\\ \\
\text{[Clarifying Question]}\\
\{lawyer question\}\\ \\
\text{[User Response]}\\
\{User Response\}\\ \\
}\normalsize
\end{tcolorbox}

\subsection{Evaluation Results for User Agent}

As shown in Table \ref{tbl:user relevance}, Qwen-max with First-Person Given Dialogue approach achieved outstanding performance in both relevance (97.0\%) and factuality (99.3\%) metrics.
This superior performance demonstrates the effectiveness of using first-person perspective in user simulation tasks. 
Notably, when comparing different model scales, we also observe a clear scaling law pattern: larger LLMs consistently outperform their smaller counterparts across all evaluation settings. 
For instance, Qwen2.5-32B shows improvements over Qwen2.5-7B, and Qwen2.5-72B further enhances the performance.

\begin{table*}[h]
\vspace{-3mm}
  \caption{ Results on User Reliable.}
  \label{tbl:user relevance}
  \centering
  \scalebox{0.85}{
  \begin{tabular}{l|c|c|c}
    \toprule
    Models&Instruct Given Dialogue&Instruct Given Question&First-Person Given Dialogue\\
    \hline
 \multicolumn{4}{c}{Relevance}\\
     \hline
 Qwen2.5-7B&86.1& 90.6& 86.5
\\
 Qwen2.5-32B&90.2& 91.1& 96.9
\\
 Qwen2.5-72B&90.9& 90.0& 93.7
\\
 Qwen-max& 92.4& 92.2&\textbf{97.0}\\
     \hline

 \multicolumn{4}{c}{Factuality}\\
     \hline
 Qwen2.5-7B& 85.0& 94.3&96.0\\
 Qwen2.5-32B& 94.5& 98.0&99.5\\
 Qwen2.5-72B& 97.5& 98.0&99.5\\
     Qwen-max&99.0&99.0& \textbf{99.3}\\
    \bottomrule
  \end{tabular}
  }
  \vspace{-3mm}
\end{table*}

Interestingly, even smaller models like Qwen2.5-7B demonstrate reasonable performance (86.5\% relevance and 96.0\% factuality).
This robust performance across different model scales indicates that the user simulation approach is feasible and practical even with more accessible open-source models.
The high performance across different model sizes validates the viability of using LLMs for user simulation in legal consultation scenarios, making it a promising direction for developing and testing legal consultation systems.

\section{Lawyer Agent}
\label{app:lawyeragent}
\subsection{Experiment Setting}
\label{app:experiment setting}

\subsubsection{Experiment Setting for API-Based LLMs}

We first present the experimental settings for LLMs accessed via API calls, including closed-source models, open-source models, and several legal domain-specific models. These models are evaluated using both zero-shot and few-shot strategies with no additional fine-tuning.
All prompts for zero-shot and few-shot settings can be found in Appendix \ref{app:zeroshot}.

First, we introduce the versions of each general LLM (include both open-source and close-source LLMs) as shown in Table \ref{tbl:fs_mainresult1}.
For closed-source LLMs (GPT-series, qwen-max, qwen-turbo, farui) and deepseek-series (deepseek-v3, and deepseek-r1), we directly call APIs for generation.
For all other models, the evaluation is conducted using vllm~\citep{vllm} to enable an OpenAI-Compatible Server for model calls, utilizing 4 NVIDIA A100 GPUs with 80GB of memory.
The maximum sequence token is set to 2048 and temperature for generation is set to 0.7 for open-source models.

\begin{table*}[ht]
  \caption{LLMs version and source url.}
  \label{tbl:fs_mainresult1}
  \centering
  \begin{tabular}{ll}
    \toprule
    \textbf{Model Names}&  \textbf{Model Version or Source}\\
     \hline
 GPT-4&  gpt-4-turbo-2024-04-09\\
 GPT-4o&  gpt-4o-mini-2024-07-18\\
 GPT-3.5-turbo&gpt-3.5-turbo-1106\\
 Qwen-max&  qwen-max-2024-09-19\\
 Qwen-turbo&  qwen-turbo-2025-02-11\\
     \hline
    deepseek-v3&  https://huggingface.co/deepseek-ai/DeepSeek-V3\\
    deepseek-r1&  https://huggingface.co/deepseek-ai/DeepSeek-R1\\
 Qwen2.5-72B&  https://huggingface.co/Qwen/Qwen2.5-72B-Instruct\\
 Qwen2.5-7B&  https://huggingface.co/Qwen/Qwen2.5-7B-Instruct\\
 Llama-3.1-8B&  https://huggingface.co/meta-llama/Llama-3.1-8B-Instruct\\
 GLM4-32B&  https://huggingface.co/THUDM/GLM-4-32B-0414\\
 GLM4-9B&  https://huggingface.co/THUDM/glm-4-9b-chat\\
     \hline
 ChatLaw&  https://huggingface.co/pandalla/ChatLaw-13B\\
 Lawyer-LLaMA&  https://github.com/AndrewZhe/lawyer-llama\\
     farui-plus& https://help.aliyun.com/zh/model-studio/tongyi-farui-api\\
    \bottomrule
  \end{tabular}
\end{table*}

\subsubsection{Experiment setting for Mediq Strategy}

We adapt the mediq-expert framework~\citep{li2024mediq}, originally designed for clinical decision-making through information-seeking questions, to our legal consultation scenario. While the original framework was developed for medical multiple-choice questions, we modify and retain three key components: the Abstention module, Ask module, and a Suggestion module for legal advice generation.

The workflow operates as follows: The Abstention module first assesses the system's confidence level for providing legal advice. 
When confidence is low, the Ask module activates to gather additional information through asking a clarifying question.
The dialogue concludes either when the system reaches sufficient confidence or when the maximum number of interaction turns is reached, at which point the Suggestion module generates appropriate legal advice.

We utilize Qwen-max as backbone LLM for each module in mediq.
All prompts for mediq-expert can be found in Appendix \ref{app:mediq}.

\subsubsection{Experiment setting for SFT Strategy}
\label{app:sftstrategy}
To enhance LLMs' legal consultation capabilities, we introduce several strategies to construct training multi-turn dialogue for effectively training LLMs by 
Supervised Fine-Tuning (SFT).

\textbf{Direct SFT} directly leverages the raw dialogues of LeCoDe training data to predict lawyer utterances based on dialogue history, simulating natural consultation flow.
Formally, given a consultation dialogue sequence $D=\{(u_t^C,u_t^E)\}_{t=1}^T$, where $u_t^C$ and $u_t^E$ represent utterances of Client $C$ and Expert (lawyer) $E$ at turn $t$,
the model learns to generate lawyer responses sequentially. For instance, when a client initiates with query $u_1^C$, the model learns to predict the lawyer's first response $u_1^E$.
Subsequently, using the concatenated context ($u_1^C,u_1^E,u_2^C$), the model learns to predict the next lawyer utterance $u_2^E$.

\textbf{Key-fact SFT} employs a targeted approach where stronger LLMs first generate multiple target questions for each atomic key fact. The target model is then trained to learn how to generate these target questions, establishing a one-to-one mapping between questions and facts.
Formally, given each key fact $a_t$ in the atomic key facts $A_n$, we use Qwen-max to generate a target question designed to elicit specific information $a_t$.
Then we construct the new multi-turn dialogue sequence $D=\{(q_t,a_t)\}_{t=1}^N$, where $q_t$ represents generated target question and $a_t$ represents corresponding atomic key fact.
The model learns to generate appropriate questions sequentially.

For instance, when given an initial query, the model learns to generate $q_1$ to extract $a_1$. Subsequently, using the concatenated context $(q_1,a_1)$, it predicts the next question $q_2$ to obtain $a_2$, ensuring systematic information gathering through targeted questioning.
Following this strategy, we can construct the Key-fact SFT training dialogue.

\textbf{Key-fact-e SFT}, an enhanced version of the previous approach, also utilizes stronger LLMs (qwen-max) to generate target questions that can effectively cover multiple (1-3) key facts simultaneously. Training data is then constructed based on these comprehensive questions and their corresponding key facts, enabling more efficient information elicitation.

For all strategies,
we fine-tune Qwen2.5-7B-Instruct using the same prompt template from zero-shot experiments in Appendix \ref{app:zeroshot}. 
The prompt templates for Qwen-max to generate the target questions in Key-fact SFT and Key-fact-e SFT are provided in Appendix \ref{app:sft_strategies_prompt}.
We run experiments on 4 NVIDIA A100 GPUs with 80GB of memory.
The training configuration employs LoRA (rank=8, alpha=16) with an effective batch size of 64 (4 samples per device × 4 gradient accumulation steps). The optimization process uses a cosine learning rate schedule initialized at 1e-5 with 10\% warmup ratio over 3 epochs.

\subsection{Lawyer Agent Prompt}

\subsubsection{Zero-shot Prompt}
\label{app:zeroshot}

\begin{tcolorbox}[title={Lawyer Agent Zero-shot Prompt for ask or advice (translated from Chinese)}, colframe=gray]
\small\texttt{\textcolor{teal}{You are a lawyer well-versed in Chinese law, responsible for providing legal consultation to users. Your task is to ask clarifying questions to better understand the details of the user's case or providing legal advice.}\vspace{4mm}\\
Users may present unclear or non-professional questions. In such cases, ensure your questions directly target key information, guiding users to provide more relevant details. The goal is to make users' needs clearer to provide more accurate legal advice.\\ \\
Below is the user's query and your interaction dialogue:\\
\{current dialogue\}\\ \\
Notes:\\
a. You have two operations: 1. Question, 2. Advice. Choose one operation per round.\\
b. If user information is insufficient, choose operation 1. Question.\\
-- Ask only one critical question per round.\\
-- Questions should be concise, without any additional information.\\
-- Do not ask repeated questions!!\\
-- Questions should verify specific facts. Considering users seek legal help but lack legal background, avoid asking about typical court sentencing or specific law violations.\\
-- Maximum 10 rounds of questions; must provide advice after 10 rounds.\\
c. When confident in addressing user needs, choose operation 2. Advice to provide professional legal advice.\\
d. Response template:\\
``question:a clarification question'' or ``advice:legal advice''
}\normalsize
\end{tcolorbox}

\begin{tcolorbox}[title={Lawyer Agent Zero-shot Prompt for advice generation (translated from Chinese)}, colframe=gray]
\small\texttt{\textcolor{teal}{You are a lawyer well-versed in Chinese law, responsible for providing legal consultation to users. Your task is to provide legal advice.}\vspace{4mm}\\
Below is the user's query and your interaction dialogue:\\
\{current dialogue\}\\ \\
Notes:\\
a. Your task is to provide legal advice according to the context.\\
b. Response template:\\
``advice:legal advice''
}\normalsize
\end{tcolorbox}

\subsubsection{Few-shot Prompt}
\label{app:fewshot}

\begin{tcolorbox}[title={Lawyer Agent Few-shot Prompt for ask or advice (translated from Chinese)}, colframe=gray]
\small\texttt{\textcolor{teal}{You are a lawyer well-versed in Chinese law, responsible for providing legal consultation to users. Your task is to ask clarifying questions to better understand the details of the user's case or providing legal advice.}\vspace{4mm}\\
Users may present unclear or non-professional questions. In such cases, ensure your questions directly target key information, guiding users to provide more relevant details. The goal is to make users' needs clearer to provide more accurate legal advice.\\ \\
Below is two demonstrations:\\
<demo1>\\
\{demo1\}\\ \\
<demo2>\\
\{demo2\}\\ \\
\{current dialogue\}\\ \\
Below is the user's query and your interaction dialogue:\\
\{current dialogue\}\\ \\
Notes:\\
a. You have two operations: 1. Question, 2. Advice. Choose one operation per round.\\
b. If user information is insufficient, choose operation 1. Question.\\
-- Ask only one critical question per round.\\
-- Questions should be concise, without any additional information.\\
-- Do not ask repeated questions!!\\
-- Questions should verify specific facts. Considering users seek legal help but lack legal background, avoid asking about typical court sentencing or specific law violations.\\
-- Maximum 10 rounds of questions; must provide advice after 10 rounds.\\
c. When confident in addressing user needs, choose operation 2. Advice to provide professional legal advice.\\
d. Response template:\\
``question:a clarification question'' or ``advice:legal advice''
}\normalsize
\end{tcolorbox}

\begin{tcolorbox}[title={Lawyer Agent Few-shot Prompt for advice generation (translated from Chinese)}, colframe=gray]
\small\texttt{\textcolor{teal}{You are a lawyer well-versed in Chinese law, responsible for providing legal consultation to users. Your task is to provide legal advice.}\vspace{4mm}\\
Below is two demonstrations:\\
<demo1>\\
\{demo1\}\\ \\
<demo2>\\
\{demo2\}\\ \\
Below is the user's query and your interaction dialogue:\\
\{current dialogue\}\\ \\
Notes:\\
a. Your task is to provide legal advice according to the context.\\
b. Response template:\\
``advice:legal advice''
}\normalsize
\end{tcolorbox}

\begin{tcolorbox}[title={Demonstration 1 (translated from Chinese)}, colframe=gray]
\small\texttt{<demo1>\\
User: On August 21st, I heard he was suspected of fraud. When police went to arrest him, he injured three people while fleeing but was eventually caught.\\
Lawyer: Question: Could you first clarify if the suspect was alone at the time?\\
User: His girlfriend was with him and also got arrested, but she was released that night.\\
Lawyer: Question: Do you know the specific amount involved in the alleged fraud?\\
User: I don't know the exact amount as we can't contact the suspect now.
Lawyer: Question: Where did the incident take place?\\
User: It happened in a public place in another city.\\
Lawyer: Question: Did he intentionally crash into the crowd?\\
User: Not sure if he deliberately rushed toward the crowd.\\
Lawyer: Advice: First, we need to arrange a lawyer to meet with your cousin to understand the details of the fraud allegations and the injuries of the three people. Currently, we can't determine the specific sentence as it depends on the severity of injuries and specific circumstances. If the injuries are minor and only involved three people, it might be classified as assaulting police or obstructing official duties; if there were many people present and he was driving recklessly, it could constitute endangering public safety by dangerous means, which carries more severe penalties. After understanding all relevant information, we can develop a targeted strategy. Additionally, trying to obtain forgiveness letters from the three police officers is an important step, but this depends on their willingness. In summary, gathering all relevant information is the crucial first step.\\
}\normalsize
\end{tcolorbox}

\begin{tcolorbox}[title={Demonstration 2 (translated from Chinese)}, colframe=gray]
\small\texttt{<demo2>\\
User: My husband has several mistresses. \\
Lawyer: Question: What's your husband's occupation? Where does he get so much money? \\
User: He's a procurement officer, responsible for company purchases. He accepts kickbacks from suppliers, mostly illegal gains. \\
Lawyer: Question: Do you own any properties? Are you aware of the source of funds for these properties? \\
User: I own two houses, but I suspect the funds might be related to his illegal income. \\
Lawyer: Question: Have you previously discussed divorce with your husband? Were any specific agreements reached? \\
User: We negotiated divorce terms, demanding 200,000 payment, with a 400,000 penalty clause, but important terms were deleted and no agreement was reached. \\
Lawyer: Question: What are your current preliminary arrangements regarding property division and child support? \\
User: Currently, both houses belong to the client, with 5,000 monthly child support. Husband will give client an additional 200,000. Planning to draft a new divorce agreement and consider further legal action against husband after everything is settled. \\
Lawyer: Advice: Non-state personnel bribery of over 30,000 yuan constitutes a crime. In handling this matter, you can negotiate with him, clearly state the evidence you have, and present your conditions, but immediate legal action isn't necessary. If you decide to take further steps, prepare materials like a PPT presentation and send the cover first, indicating you've started collecting evidence for reporting to authorities. Regarding the divorce agreement, detail specific child support items (like living expenses, education fees, and major medical expenses), and set the 200,000 compensation payment for one or two years later to ensure you first secure major assets like the properties. Considering the child's future interests, although legal action is possible, maintaining his employment might be more beneficial for long-term financial support. Regarding the mistresses, if funds are illegal, legal action could be considered, but proceed cautiously to avoid affecting core objectives.
}\normalsize
\end{tcolorbox}

\subsubsection{Mediq Prompt}
\label{app:mediq}

\begin{tcolorbox}[title={Mediq: Abstain Module(translated from Chinese)}, colframe=gray]
\small\texttt{\textcolor{teal}{You are a lawyer well-versed in Chinese law, responsible for providing legal consultation to users. Your task is to provide legal advice.}\vspace{4mm}\\ \\
Below is the user's query and your interaction dialogue:\\
\{current dialogue\}\\ \\
Evaluating above case facts, are you confident to provide legal advice according to the current dialogue?\\
Answer with YES or NO and NOTHING ELSE.
}\normalsize
\end{tcolorbox}

\begin{tcolorbox}[title={Mediq: Ask Module(translated from Chinese)}, colframe=gray]
\small\texttt{\textcolor{teal}{You are a lawyer well-versed in Chinese law, responsible for providing legal consultation to users. Your task is to provide legal advice.}\vspace{4mm}\\ \\
Your task is to ask clarifying questions to better understand the details of the user's case or providing legal advice.\\
Below is the user's query and your interaction dialogue:\\
\{current dialogue\}\\ \\
Notes:\\
a. You have ask a clarifying question.\\
b. If user information is insufficient, choose operation 1. Question.\\
-- Ask only one critical question per round.\\
-- Questions should be concise, without any additional information.\\
-- Do not ask repeated questions!!\\
-- Questions should verify specific facts. Considering users seek legal help but lack legal background, avoid asking about typical court sentencing or specific law violations.\\
-- Maximum 10 rounds of questions; must provide advice after 10 rounds.\\
b. Response template:\\
``question:a clarification question''
}\normalsize
\end{tcolorbox}

\begin{tcolorbox}[title={Mediq: Suggestion Module(translated from Chinese)}, colframe=gray]
\small\texttt{\textcolor{teal}{You are a lawyer well-versed in Chinese law, responsible for providing legal consultation to users. Your task is to provide legal advice.}\vspace{4mm}\\
Below is the user's query and your interaction dialogue:\\
\{current dialogue\}\\ \\
Notes:\\
a. Your task is to provide legal advice according to the context.\\
b. Response template:\\
``advice:legal advice''
}\normalsize
\end{tcolorbox}

\subsubsection{SFT Strategies Prompt}
\label{app:sft_strategies_prompt}

\begin{tcolorbox}[title={Key-Fact SFT to generate a target question (translated from Chinese)}, colframe=gray]
\small\texttt{\textcolor{teal}{You are a lawyer well-versed in Chinese law, responsible for providing legal consultation to users. }\vspace{4mm}\\ \\
I will provide a list of key information points that need to be obtained from clients but are currently unknown. Please design professional, guiding questions for each information point to effectively elicit these details.\\
\text{[Atomic Key Facts]}\\
\{Atomic Key Facts\}\\ \\
Requirements:\\
1. Return a dictionary mapping each key information index to a corresponding question\\
2. Each question should:\\
-Use neutral, professional terminology\\
-Avoid leading language\\
-Be easily understood by clients\\
-Naturally elicit the target information\\
-Be concise\\}\normalsize
\end{tcolorbox}

\begin{tcolorbox}[title={Key-Fact-e SFT to generate target questions (translated from Chinese)}, colframe=gray]
\small\texttt{\textcolor{teal}{You are a lawyer well-versed in Chinese law, responsible for providing legal consultation to users. }\vspace{4mm}\\ \\
I will provide a list of key information points that need to be obtained from clients but are currently unknown. Please design a series of professional, guiding questions that efficiently cover these information points.\\
\text{[Atomic Key Facts]}\\
\{Atomic Key Facts\}\\ \\
Requirements:\\
1. Return a dictionary where:\\
- Each key is a question
- Each value is a list of indices indicating which key information points the question covers
2. Each question should:\\
-Use neutral, professional terminology\\
-Avoid leading language\\
-Be easily understood by clients\\
-Naturally elicit the target information\\
-Be concise\\}\normalsize
\end{tcolorbox}

\section{Details of Annotation Process}
\label{appendix:annotation}

\subsection{Guidelines for Annotation Stage-1}

This annotation process consists of two steps: (1) Dialogue Standardization: quality checking of legal consultation dialogues to correctly identify speaker roles, correct typos, and identify coded language; (2) Dialogue Intent Annotation: labeling dialogue intents.

\textbf{1. Dialogue Standardization}
The dialogues are from live legal consultation scenarios, involving conversations between (1) clients and (2) lawyers. Clients present legal inquiries, and lawyers address these through questioning, confirmation, and advice. Since the data is transcribed by large language models, there may be errors in role assignments and speech recognition. This phase aims to check and correct dialogue quality.

Specific steps include:

1. Verify correct role assignment between lawyer and client:
   \begin{itemize}
   \item Case 1: All role assignments are incorrect - Fix all roles accordingly
   \item Case 2: Individual statement roles are incorrect - Identify and adjust misattributed statements
   \item Case 3: Single statements containing multiple speakers - Separate into distinct turns
   \end{itemize}

2. Jargon and Euphemism Detection: Identify and convert substitute terms used to bypass platform filters into standard text.
   Example: "hat uncle" → "police officer"
   Note: All potential euphemisms and substitute terms will be listed for further assessment.

3. Correct transcription errors, including typos and homophone mistakes.

4. Ensure dialogue format:
   \begin{itemize}
   \item Combine consecutive statements by the same speaker
   \item Ensure the dialogue begins with client's initial query and ends with lawyer's advice
   \item Remove greeting/farewell phrases like "hello," "thanks," "goodbye"
   \end{itemize}

\textbf{2. Dialogue Intent Annotation}
Based on quality-checked dialogue data, label each utterance with one of these intent categories:

\begin{enumerate}
\item "Client Initial Query": Client's opening question (typically first statement)
\item "Client Response": Client's answers to lawyer's clarifying questions
\item "Client Information Addition": Client's voluntary provision of additional key information
\item "Client Need Extension": Client's new questions/needs arising during consultation
\item "Lawyer Clarifying Question": Lawyer's clarifying questions to gather more case information
\item "Lawyer Information Verification": Lawyer's repeated confirmation of crucial points
\item "Lawyer Legal Advice": Lawyer's legal recommendations based on case understanding
\item "Lawyer Emotional Support": Lawyer's emotional encouragement or comfort to client
\item "Invalid Exchange": Irrelevant dialogue not affecting final legal advice
\item "Other": Miscellaneous categories not covered above
\end{enumerate}

\subsection{Guidelines for Annotation Stage-2}

Stage 2 annotation primarily involves (1) labeling atomic key facts information and (2) summarizing lawyer's advice.

\textbf{1. Atomic Key Facts Information Labeling}
Since clients' initial inquiries often lack complete information, based on the Q\&A process between clients and lawyers, Qwen-max has preliminarily extracted Atomic Key Facts List.
Further extraction and verification are needed to compile a complete list of critical information that informed the lawyer's legal advice.

Requirements:
\begin{enumerate}
\item Extract key information points from client-lawyer dialogue. Keep expressions consistent with original text where possible.

\item Cross-check Atomic Key Facts List against dialogue points for completeness. Add missing information as needed.

\item Merge similar key fact information points to avoid redundancy. Atomic Key Fact Points should be atomic-level:

Example:
\begin{quote}
- ["Client", "They had a banquet that evening.", "Client Response"],

- ["Lawyer", "What kind of banquet at friend A's house?", "Lawyer Clarifying Question"],

- ["Client", "It was their child's first birthday.", "Client Response"],

- ["Lawyer", "First birthday banquet. I see.", "Invalid Exchange"],

- ["Client", "Then got very drunk that night.", "Client Information Addition"],

- ["Lawyer", "A got drunk.", "Lawyer Information Verification"],

- ["Client", "Yes, intoxicated.", "Client Response"]

Corresponding key fact information point:\

"Friend A hosted child's first birthday banquet that evening, A got heavily intoxicated"
\end{quote}

\item Supplement missing entity information (e.g., specific roles or event descriptions).\
Example: Change "was detained" to "Defendant A was detained"

\item Maintain chronological order of atomic facts consistent with dialogue flow.

\item Ensure extracted information comes from client statements only, excluding lawyer's advice.
However, combine lawyer questions with client answers when relevant.

\item Rate importance of each atomic fact:
\begin{itemize}
\item Critical Facts (3 points): Directly affects case classification, liability distribution, or rebuts prosecution's charges
\item Secondary Facts (2 points): Provides background or supporting information with moderate case impact
\item Non-critical Facts (1 point): Subjective opinions or procedural descriptions with no direct impact on fact-finding
\end{itemize}
\end{enumerate}

\textbf{2. Lawyer Advice Summary}
Legal advice may be scattered throughout the dialogue. Consolidate all advice while:
\begin{itemize}
\item Filtering out meaningless information
\item Preserving lawyer's original key terminology to maintain accuracy and completeness
\item Avoiding alterations to lawyer's original intent and critical information
\end{itemize}

\section{Dataset}

\subsection{Sample Data Format}
\label{appendix:subsection sampledata}
Below is an example of our annotated legal consultation data structure:

\begin{verbatim}
{
    'id': 'xxx',
    'dialogue': [
        ['user', "xxx"],
        ['lawyer', "xxx"],
        ...
    ],
    'initial_query': 'X',
    'suggestion': 'suggestion.',
    'key_fact_dict_hard_version': {
        'atomic_facts': {
            '0': 'a',
            '1': 'B',
            '2': 'c',
            '3': 'D',
            '4': 'e',
            '5': 'f',
            '6': 'xx',
            '7': 'xxx',
            '8': 'xxx',
            '9': 'xx'
        },
        'importance_scores': {
            '0': 2.0,  # Secondary fact
            '1': 1.0,  # Non-critical fact
            '2': 3.0,  # Critical fact
            '3': 2.0,  # Secondary fact
            '4': 3.0,  # Critical fact
            '5': 3.0,  # Critical fact
            '6': 3.0,  # Critical fact
            '7': 2.0,  # Secondary fact
            '8': 3.0,  # Critical fact
            '9': 1.0   # Non-critical fact
        }
    },
    'case_type': ['Criminal'], # ['Criminal','Civil','Administative']
    'charge_type': ['Intentional Homicide'],
    'intention_annotation': [
        ['Client', 
         'B was working as a hotel receptionist when B, C, and D were forcibly 
          taken into a room.',
         '[1] Client Initial Query'],
        ...
    ]
}
\end{verbatim}

Note: Names have been anonymized (A, B, C, etc.) for privacy protection. The importance scores are categorized as: 3.0 (Critical fact), 2.0 (Secondary fact), and 1.0 (Non-critical fact).

\subsection{Evaluation Metrics}
\label{app:evaluation metrics}
We evaluate expert responses from two main perspectives. 
First, we assess \textbf{Clarification Capability} through \textit{effectiveness} and \textit{efficiency} metrics.
Second, we evaluate \textbf{Advice Quality} using both \textit{automated} and \textit{LLM-based evaluation} metrics.

\subsubsection{Clarification Capability}

\textbf{Effectiveness}

Given ground truth atomic key facts list $A_n$ and simulation dialogue, we use qwen-max to match how many key facts from $A_n$ are mentioned in the user-lawyer consultation dialogue. The extraction prompt template is shown in Table \ref{tbl:extractkeyfact}.

We designed four metrics to evaluate clarification capability:

1. \textbf{Recall (Rec.)}: measures the comprehensiveness of key fact coverage, defined as:

\begin{equation}
\text{Rec.} = \frac{1}{|N|} \sum_{i=1}^{N} \frac{|\mathcal{P}_i \cap A_i|}{|A_i|}
\end{equation}

where $\mathcal{P}_i$ represents the predicted key facts set for the $i$-th dialogue, $A_i$ denotes the corresponding ground truth set, and $N$ is the total number of dialogues in the evaluation set.

2. \textbf{Weighted Recall (Weighted Rec.)}: incorporates the significance of each key fact through importance weights, defined as:

\begin{equation}
\text{Weighted Rec.} = \frac{1}{|N|} \sum_{i=1}^{N} \frac{\sum_{j \in \mathcal{P}_i \cap A_i} w_j}{\sum_{k \in A_i} w_k}
\end{equation}

where $w_j \in \{1,2,3\}$ denotes the importance weight assigned to each key fact $j$, $\mathcal{P}_i$ represents the predicted key facts set for the $i$-th dialogue, and $A_i$ denotes the corresponding ground truth set. The weights reflect the relative significance of each key fact in the legal consultation context.

3. \textbf{Recall@5 (Rec@5)}: evaluates how effectively the first five questions elicit key facts, defined as:

\begin{equation}
\text{R@5} = \frac{1}{|N|} \sum_{i=1}^{N} \frac{|\mathcal{P}_i^5 \cap A_i|}{|A_i|}
\end{equation}

where $\mathcal{P}_i^5$ represents the set of key facts identified in the first five questions of the $i$-th dialogue, and $A_i$ denotes the complete ground truth set. This metric specifically assesses the model's ability to efficiently extract crucial information in the early stages of the consultation.

4.\textbf{ Normalized Discounted Cumulative Gain (NDCG)}: evaluates the effectiveness of question ordering by considering both the importance of key facts and their position in the dialogue sequence, defined as:

\begin{equation}
\text{NDCG} = \frac{\text{DCG}}{\text{IDCG}}
\end{equation}

where DCG (Discounted Cumulative Gain) is calculated as:

\begin{equation}
\text{DCG} = \sum_{i=1}^{n} \frac{rel_i}{\log_2(i+1)}
\end{equation}

and IDCG (Ideal DCG) is:

\begin{equation}
\text{IDCG} = \sum_{i=1}^{n} \frac{rel_i^*}{\log_2(i+1)}
\end{equation}

Here, $rel_i$ represents the importance score of a key fact discovered at position $i$, and $rel_i^*$ denotes the importance scores sorted in descending order for the ideal sequence. Each key fact is counted only once at its first appearance in the dialogue, with importance scores ranging from 1 to 3.

\textbf{Efficiency}

We measure the average number of dialogue turn rounds (\textbf{AVG T}).

\begin{equation}
\text{AVG T} = \frac{1}{|N|} \sum_{i=1}^{N} |T_i|
\end{equation}

where $|T_i|$ represents the number of turns in the $i$-th dialogue, and $N$ is the total number of dialogues in the evaluation set. A turn $T$ is defined as one complete question-answer pair between the lawyer and user, excluding the initial user query and the final legal advice. This metric specifically measures the efficiency of the lawyer's clarification process during the consultation.

\subsubsection{Advice Quality}

We evaluate the semantic alignment between the generated and reference advice through both \textit{automated metrics} and \textit{LLM-based evaluation}. 

\paragraph{Automated Metrics} We employ two standard metrics:

\begin{itemize}
    \item \textbf{ROUGE-L}: We calculate ROUGE-L~\citep{lin-2004-rouge} between reference and generated advice after Chinese word segmentation using jieba package~\footnote{https://pypi.org/project/jieba/}.
    \item \textbf{BERTScore}: Leverages BERT contextual embeddings to compute semantic similarity~\citep{bertscore}.
\end{itemize}

\paragraph{LLM-based Evaluation} We assess five dimensions on a scale of 1-10:
\begin{itemize}
    \item \textbf{Professionalism (Pro.)}: 
        \begin{itemize}
            \item Accurate understanding and relevant solutions
            \item Clear explanation of complex legal concepts
            \item Actionable recommendations
        \end{itemize}
    
    \item \textbf{Fluency (Flu.)}: 
        \begin{itemize}
            \item Semantic coherence without logical errors
            \item Consistency in style and content
            \item Friendly and engaging response tone
        \end{itemize}
        
    \item \textbf{Completeness (Com.)}: 
        \begin{itemize}
            \item Sufficient information and details
            \item Coverage of essential recommendations
        \end{itemize}
        
    \item \textbf{Satisfaction (Sat.)}: 
        \begin{itemize}
            \item Targeted and personalized solutions
            \item Accessible language and expression
            \item Empathy and respect for client concerns
        \end{itemize}
        
    \item \textbf{Safety (Safe.)}: 
        \begin{itemize}
            \item Scientific and accurate legal knowledge
            \item Prevention of potentially harmful advice
            \item Adherence to professional ethics and non-discriminatory content
        \end{itemize}
\end{itemize}

These five dimensions are aggregated into an \textbf{Overall Score (OA)} on a scale of 1-10, reflecting the comprehensive quality of the legal advice.

We employ GPT-4-mini as an automated judge to evaluate the generated legal advice across five dimensions (1-10 scale).
The specific prompting strategy for LLM-as-a-judge evaluation is detailed in Table~\ref{tbl:advicerating template}.

\section{More Experimental Results}

\begin{table*}[ht]
  \caption{The results on Clarification Capability, comparing few-shot (FS) and zero-shot (ZS) strategies.}
  \label{tbl:app_fs_mainresult1}
  \centering
  \begin{tabular}{lc|cccc|c}
    \toprule
    \multirow{2}{*}{\textbf{Models}}&  &\multicolumn{4}{c}{\textbf{Effectiveness}}& \textbf{Efficiency} \\
 &  &Rec.(\%)& Weighted Rec.(\%)& Rec@5(\%)& NDCG(\%)& Avg T $\downarrow$\\
    \midrule
      \multicolumn{7}{c}{\textit{Closed-source LLMs}}\\
     \hline
 GPT-4&  ZS&35.9& 37.2& 34.4& 74.7&4.9\\
     &  FS&32.5& 33.6& 31.9& 71.4&4.3\\
 GPT-4o&  ZS&33.0& 33.9& 25.9& 63.2& 10\\
 &  FS&33.6& 34.2& 27.9& 65.6&10\\
 Qwen-max&  ZS&39.8& 41.3& 31.2& 70.7& 10\\
 &  FS&40.0& 41.2& 30.8& 70.5& 10\\
 Qwen-turbo&  ZS&27.3& 28.3& 23.2& 61.2& 9.4\\
 &  FS&25.7& 26.8& 22.7& 63.5&9.8\\
 \hline
       \multicolumn{7}{c}{\textit{Open-source LLMs}}\\
     \hline
    deepseek-v3&  ZS&28.0&29.0& 25.1& 65.4&  7.5\\
 &  FS&32.3& 33.4& 27.1& 65.3&9.1\\
    deepseek-r1&  ZS&27.6&28.8& 25.1& 63.9&  6.3\\
 &  FS&32.0& 33.2& 26.2& 64.1&8.4\\
 Qwen2.5-72B&  ZS&39.6& 40.8& 30.4& 71.9& 10\\
 &  FS&17.2& 17.7& 13.3& 33.7&7.6\\
 Qwen2.5-7B&  ZS&31.1& 32.1& 27.7& 70.1& 10\\
 &  FS&24.7& 25.4& 22.7& 64.1&10\\
 Llama-3.1-8B&  ZS&22.1& 23.1& 19.1& 55& 7.8\\
 &  FS&21.3& 22& 18.7& 56.4&9.6\\
 GLM4-32B&  ZS&34.1& 35.4& 28.8& 67.3& 7.3\\
 &  FS&38.5& 39.8& 30.1& 68.2&8.6\\
 GLM4-9B&  ZS&34.2& 35.3& 26.1& 65.0& 10\\
 &  FS&30.9& 31.9& 24.9& 64.0&10\\
 \hline
       \multicolumn{7}{c}{\textit{Domain-specific LLMs or Clarifying Strategy}}\\
     \hline
 ChatLaw&  ZS&16.7& 17.4& 16.3& 47.8& 4.4\\
 Lawyer-LLaMA&  ZS&21.8& 22.9& 21.8& 62.9& 4.2\\
  Farui&  ZS&28.4& 29.5& 25.1& 64.2& 8.9\\
 MediQ&  ZS&22.1& 22.9& 19.0& 53.5& 4.2\\
 Direct SFT&  ZS&37.2& 37.6& 25.6& 62.3& 9.6\\
 Key-fact SFT& ZS& 53.8& 55.1& 43.5& 84.7&9.2\\
     Key-fact-e SFT& ZS&51.0&51.1& 45.3& 84.8&  6.6\\
    \bottomrule
  \end{tabular}
\end{table*}

\begin{table}[h]
  \caption{The results on Advice Quality, comparing few-shot (FS) and zero-shot (ZS) strategies.}
  \label{tbl:app_fs_mainresult2}
  \centering
  \begin{tabular}{l|lcc|cccccc}
    \toprule
    \multirow{2}{*}{\textbf{Models}} && \multicolumn{2}{c}{\textbf{Automated Metrics}}&\multicolumn{6}{c}{\textbf{LLM-Evaluaion Metrics}}\\
  && Rouge-L& BERTSscore& Pro.& Flu.& Com.& Sat.& Safe.&OA\\
    \midrule
     \multicolumn{10}{c}{\textit{Closed-source LLMs}}\\
     \hline
 GPT-4 &ZS& 17.5& 65.9& 56.8& 71.9& 49.9& 55.8& 71.1&58.1\\
  &FS& 17.4& 66.1& 59.2& 72.5& 52.5& 58.2& 72.1&59.9\\
 GPT-4o &ZS& 13.0& 62.9& 50.1& 69.6& 44.7& 52.4& 66.6&53.5\\
  &FS& 16& 64.6& 54.8& 72.7& 49.4& 56.3& 69.9&57.5\\
     Qwen-max &ZS& 13.9& 63.9& 54.0& 71.4& 47.6& 55.3& 68.9&56.2\\
  &FS& 16.1& 65.4& 56.5& 72.9& 50.6& 56.9& 71.7&58.5\\
    Qwen-turbo &ZS& 10.6& 61.8& 44.8& 62.7& 40.0& 47.6& 61.6&48.5\\
  &FS& 13.5& 63.6& 53.3& 70.1& 47.1& 53.4& 69&55.3\\
    \hline
 \multicolumn{10}{c}{\textit{Open-source LLMs}}\\
 \hline
 deepseek-v3 &ZS& 16.2& 65.3& 57.6& 72.9& 50.8& 56.2& 72.1&58.9\\
   &FS& 16.5& 65.6& 58.7& 73.7& 51.9& 57.5& 72.9&59.7\\
 deepseek-r1 &ZS& 12.4& 64& 63.8& 74.1& 55.5& 55.9& 76.5&62.2\\
   &FS& 13.4& 64.8& 66.7& 75& 58.8& 58.2& 78.3&64.6\\
 Qwen2.5-72B &ZS& 14.7& 63.8& 54& 71.6& 48.5& 55.2& 69.4&56.6\\
   &FS& 10.1& 57.7& 33.6& 42.7& 30.4& 33.8& 42.5&34.7\\
 Qwen2.5-7B &ZS& 9.2& 60.0& 42.3& 60.4& 37.6& 45.0& 57.8&45.7\\
   &FS& 9.2& 59.1& 40.1& 57& 34.9& 42.3& 55.3&42.9\\
 Llama-3.1-8B &ZS& 13.9& 63.1& 51.7& 69.1& 46.5& 51.7& 67.1&54.1\\
  &FS& 7.8& 54.8& 32.6& 44.2& 29.8& 34.3& 45.3&34.4\\
 GLM4-32B &ZS& 14.3& 64.1& 55.6& 70.4& 49.0& 54.3& 70.4&56.7\\
  &FS& 14.8& 64.7& 57.6& 72.4& 51& 56.1& 72.3&58.8\\
 GLM4-9B &ZS& 13.9& 63.1& 51.7& 69.1& 46.5& 51.7& 67.1&54.1\\
   &FS& 14.1& 63.1& 52.2& 69.9& 47.3& 53.1& 68&54.7\\
 \hline
 \multicolumn{10}{c}{\textit{Domain-specific LLMs or Clarifying Strategy}}\\
 \hline
 ChatLaw &ZS& 7.8& 53.0& 28.8& 40.0& 25.5& 31.0& 39.4&30.7\\
 Lawyer-LLaMA &ZS& 12.4& 59.7& 37.6& 52.7& 33.1& 39.0& 51.8&39.4\\
 Farui &ZS& 16.6& 64.5& 52.2& 66& 47.4& 51.3& 66.3&53.8\\
 MediQ &ZS& 13.3& 63.7& 51.8& 69.1& 45.5& 51.9& 66.7&54.0\\
 Direct SFT& ZS& 15.4& 64.9& 51.7& 65.3& 45.6& 50.1& 63.1&52.4\\
 Key-fact SFT&ZS& 18.5& 67.5& 59.3& 73.0& 52.3& 55.9& 71.2&59.4\\
     Key-fact-e SFT&ZS& 19.1& 67.6& 59.3& 73.6& 52.0& 56.7& 72.1&59.5\\
    \bottomrule
  \end{tabular}
\end{table}

\clearpage

\newpage

\section{Prompt Template for LeCoDe Evaluation}
\label{app:prompt_template}
\begin{table*}[ht]
\vspace{-5mm}
\caption{The Instruction Template for extraction matching key fact from simulated dialogue, to evaluate Clarification Capability (translated from Chinese).}
\begin{tabular}{ll}
\hline
\multicolumn{2}{l}{Instruction Template used to extract matching key fact from simulated dialogue.}                           \\ \hline
\multicolumn{2}{l}{\begin{tabular}[c]{@{\ }p{\textwidth}@{\ }}
Analyze key information mentioned in user-lawyer consultation dialogues and match them with a given list of Atomic Key Facts.\\ \\
Input data:\\
1. User-Lawyer Dialogue (user lawyer dialogue):\\
   - Format: {Dialogue ID: {'q': question, 'a': answer}}\\
   - Each dialogue includes lawyer's question and user's response\\ \\
2. Atomic Key Facts List (Atomic Key Facts li):\\
   - Format: {index: information content}\\
   - Index starts from 0\\ \\
Task requirements:\\
1. Identify Atomic Key Facts mentioned in the dialogues\\
2. Match identified information with items in the Atomic Key Facts list\\
3. Record the index numbers of matching information\\ \\
Output format requirements:\\
- Output should be a Python dictionary\\
- Keys: Dialogue ID (e.g., "QA0", "QA1"...)\\
- Values: List of Atomic Key Facts indices mentioned in that dialogue\\
- If no Atomic Key Facts is mentioned, return an empty dictionary {}\\
- Example format:\\
```python\\
\{"QA0":[0,1],"QA1":2,...\}\\
```\\ \\
Below are the consultation dialogue and Atomic Key Facts list:\\
$[\text{Consultation Dialogue}]$\\
{user lawyer dialogue}\\ \\
$[\text{Atomic Key Facts}]$\\
\{Atomic Key Facts\}\\
\hline
\end{tabular}} 
\end{tabular}
\label{tbl:extractkeyfact}
\end{table*}

\begin{table*}[ht]
\vspace{-5mm}
\caption{The Instruction Template for Legal Advice Quality Evaluation, to evaluate Advice Quality (translated from Chinese).}
\scalebox{0.85}{
\begin{tabular}{ll}
\hline
\multicolumn{2}{l}{Instruction Template used in  Legal Advice Quality Evaluation, to evaluate Advice Quality.}                           \\ \hline
\multicolumn{2}{l}{\begin{tabular}[c]{@{\ }p{\textwidth}@{\ }}
You are an experienced legal expert, specializing in evaluating the quality of legal advice in consultations. Please act as a fair and rigorous evaluator to objectively assess the quality of legal advice provided by the AI assistant. In your assessment, please analyze and score based on the following five key dimensions:\\ \\
1. Professionalism\\
- Accurately understand client's issues and provide relevant answers\\
- Clearly explain complex legal knowledge\\
- Actionability of advice \\
2. Fluency\\
- Ensure semantic coherence without logical errors or irrelevant information\\
- Maintain consistency in style and content\\
- Maintain friendly and enthusiastic response attitude \\
3. Completeness\\
- Provide sufficient information and details\\
- No omission of important advice \\
4. Satisfaction\\
- Targeted responses, providing personalized solutions\\
- Use approachable expressions, language affinity\\
- Focus on client experience and feelings, empathy and respect for client needs \\
5. Safety\\
- Provide scientific and accurate legal knowledge\\
- Ensure client safety; reject potentially harmful information or advice\\
- Follow professional ethics and respect client choices; must not contain discriminatory, biased, or harmful content \\ \\

We will provide the following materials: high-quality reference answers and legal advice written by the AI assistant. 
When starting the evaluation, you need to follow these reasoning steps:\\ \\
1. Compare the AI assistant's legal advice with the reference answer, point out deficiencies in the AI's answer, and explain in detail\\
2. Evaluate the AI assistant's legal advice according to the above dimensions, scoring 1-10 for each dimension\\
3. Based on the scores of each dimension, derive a comprehensive score (1-10) for the AI assistant\\
4. Your scoring should be as strict as possible and must follow these scoring rules: higher quality responses receive higher scores \\ \\

Scoring criteria:\\
1-2 points: Advice contains irrelevant content, serious errors, unverified or false information, or potentially harmful content\\
3-4 points: No major errors but clear deficiencies in legal relationship definition or key point responses, poor logical reasoning, lacks specificity, advice too general, fails to meet basic consultation requirements\\
5-6 points: Basically meets consultation requirements, accurate legal analysis but average logical completeness, addresses main points but lacks deep analysis, overall performance is moderate\\
7-8 points: Quality approaches reference answer, provides practical solutions, excellent performance in all evaluation dimensions, no obvious defects\\
9-10 points: Quality significantly exceeds reference answer, near-perfect performance in all dimensions, provides extremely valuable solutions \\ \\
Please provide detailed evaluation notes. For each dimension score, explanations must be provided. All scores should be whole numbers. Finally, return the evaluation results in the following dictionary format: \\

python\{\{'Professionalism': score, 'Fluency': score, 'Completeness': score, 'Satisfaction': score, 'Safety': score, 'Overall Score': total\}\} \\ \\

<dialogue>
\{dialogue\} </dialogue>\\ \\
<gt suggestion>\{gt suggestion\}</gt suggestion> \\ \\
<pred suggestion>\{pred suggestion\}</pred suggestion> \\ \\
Please begin the evaluation:\\
\hline
\end{tabular}} 
\end{tabular}
}
\label{tbl:advicerating template}
\end{table*}

\end{document}